\documentclass{article}
\usepackage[most]{tcolorbox}
\usepackage{amssymb}
\usepackage{amsmath}

\usepackage{algorithm}
\usepackage[noEnd=true]{algpseudocodex}

\usepackage{bm}

\usepackage{url}

\usepackage{boxedminipage}
\usepackage{graphicx}%%[draft] : do not embed figs/picts
\usepackage{float}
\usepackage{rotating}

\usepackage{xspace}
\usepackage{nicefrac}

\usepackage{comment}
\usepackage{soul}
\usepackage{verbatim}%%block comment

\usepackage{csquotes} % \textquote{}

\usepackage[T1]{fontenc}    % for accents
\usepackage[utf8]{inputenc} % for coding
\usepackage{xparse}   % \NewDocumentCommand
\usepackage{xifthen} % if-the-else see http://ctan.org/pkg/xifthen
\usepackage{pifont}
\usepackage{marvosym}

\usepackage{fullpage}
\usepackage{xstring}
\usepackage{caption}
\usepackage{subcaption}
\usepackage{csvsimple}

\usepackage{lscape}
\usepackage[colorlinks=true, linkcolor=blue, citecolor=blue]{hyperref} %% inclusion after algorithm: mandatory
\usepackage{natbib}

\newif\ifSLIDES
\SLIDESfalse

\newif\ifLONG
\LONGfalse

\newif\ifTWOCOLUMNS
\TWOCOLUMNSfalse

\input{macros-symbols.sty}
\input{macros-algorithms-code.sty}
\newcommand{\rred}[1]{{\textcolor{red}{#1}}}         %text added
\newcommand{\bblue}[1]{\textcolor{blue}{#1}}

\newcommand{\toblue}{\color{blue}\xspace}
\newcommand{\toblack}{\color{black}\xspace}

\NewDocumentCommand\paragraphmini{O{4pt} m}{%
{\vspace{#1}
\noindent{\bf #2}
}}
\NewDocumentCommand\paramini{O{4pt} m}{%
{\vspace{#1}
\noindent{\bf #2}
}}
\newcommand{\calN}{{\mathcal N}}\newcommand{\calM}{{\mathcal M}}
\newcommand{\calR}{{\mathcal R}}\newcommand{\calL}{{\mathcal L}}

\NewDocumentCommand\sbullet{g}{\noindent $\bullet$\IfNoValueTF{#1}{}{{#1}}}
\NewDocumentCommand\sbulem {g}{\noindent $\bullet$\IfNoValueTF{#1}{}{{\em #1}}}
\newtheorem{remark}{Remark}

\newcommand{\beginsupplement}{
\setcounter{table}{0}
\renewcommand{\thetable}{S\arabic{table}}%
\setcounter{figure}{0}
\renewcommand{\thefigure}{S\arabic{figure}}%

\setcounter{section}{0}
\renewcommand{\thesection}{S\Roman{section}}               %Roman numeral title
\renewcommand{\thesection}{S\arabic{section}}               %Roman numeral title
}%% end supplement

\newcommand{\beginSI}{\beginsupplement}

\input{macros-local.sty}
\title{Improved seeding strategies for \kmeans and \Kgmm} % -- Gaussian Mixture model fitting}

\author{Guillaume Carri\`ere and Fr\'ed\'eric Cazals\thanks{\ucainria}
\thanks{emails: firstname.lastname@inria.fr}}

\begin{document}
\maketitle

%% included
%% \input{seeding-main-v2.tex}

\newcommand{\cheatheight}{-.4cm}
\renewcommand{\cheatheight}{0cm}

\begin{abstract}
We revisit the randomized seeding techniques for \kmeans clustering and
\Kgmm (Gaussian Mixture model fitting with Expectation-Maximization),
formalizing their three key ingredients: the metric used for seed
sampling, the number of candidate seeds, and the metric used for seed
selection. This analysis yields novel families of initialization
methods exploiting a {\em lookahead} principle--conditioning the seed
selection to an enhanced coherence with the final metric used to
assess the algorithm, and a {\em multipass strategy} to tame down the
effect of randomization.

Experiments show a consistent constant factor improvement over
classical contenders in terms of the final metric (SSE for \kmeans,
log-likelihood for \Kgmm), at a modest overhead. In particular, for \kmeans,
our methods improve on the recently designed multi-swap strategy,
which was the first one to outperform the greedy \kmeanspp seeding.

Our experimental analysis also shed light on subtle properties
of \kmeans often overlooked, including the (lack of) correlations
between the SSE upon seeding and the final SSE, the variance reduction
phenomena observed in iterative seeding methods, and the sensitivity
of the final SSE to the pool size for greedy methods.

Practically, our most effective seeding methods are strong candidates
to become one of the--if not the--standard techniques. From a
theoretical perspective, our formalization of seeding opens the door
to a new line of analytical approaches.
\end{abstract}

%\noindent{\bf Keywords:} k-means, k-Gaussian-mixture-model-fitting, randomized seeding, clustering, Lloyd iterations.

\section{Introduction}
\label{sec:introduction}
%%i%%%%%%%%%%%%%%%%%%%%%%%%%%%%%%%%%%%%%%%%%%%%%%%%%%%%%%%%%%%%%%%%%%%%%%%%%%%%%%%

\paragraphmini{The \kmeans and \kgmm problems.}
Clustering with \kmeans and designing Gaussian mixture models (GMM) with
\kgmm algorithms play a pivotal role in unsupervised analysis.
Consider a point set $X = \{ x_1,\dots, x_n\}\subset \Rd$, to be
partitioned into a predefined number $K$ of clusters, or to be modeled 
as a mixture of $K$ multivariate Gaussian distributions.

A \kmeans clustering is a hard partition of the $n$ points into $K$
clusters, each consisting of the data points located in a Voronoi
region of the Voronoi diagram of $K$ centers -- which in general are
not data points.  The quality of the partition/clusters is assessed by
the Sum of Squared Errors (SSE) functional, namely the sum of squared
distances between a point and its nearest center.  The search space of
\kmeans is therefore the space of partitions of the point set. For
fixed $K$ and $d$, the number of partitions induced by Voronoi/power
diagrams is polynomial~\citep{inaba1994applications}. Alas, the corresponding algorithm is
not practical, which motivated the development of so-called Lloyd
iterations that iteratively update a set of initial centers~\citep{lloyd1982least}.
A related problem consists of designing a mixture of $K$ multivariate
Gaussian distributions, so as to maximize the likelihood of the
points. The search space is now the parameter space of these Gaussian
distributions, yielding a more challenging endeavor. One pivotal
technique to design such mixtures is the Expectation-Maximization
algorithm~\citep{dempster1977maximum,wu1983convergence}, an iterative
process refining an initial guess.
It may be observed that \kmeans performs a hard clustering, while
\kgmm provides a {\em responsibility} of each component for each point, which
may be seen as a soft assignment. Interestingly, \kmeans can be derived
as a limit case of EM~\citep{bishop2006pattern}.

\paragraphmini{Seeding strategies for \kmeans and \kgmm.}
Both Lloyd iterations and EM are iterative methods
heavily relying on the starting point, namely the initial centers in
\kmeans, and the initial Gaussian components in \kgmm.
In order to reduce the number of (Lloyd, EM) iterations, the init step
seeks seeds as representative as possible of the final result.
This is practically done in an iterative fashion, for $k=1,\dots,K$.
To define the $k$-th seed/component, one uses point(s) not well
described by the previously chosen/defined $k-1$ seeds/components, and
ideally provides a good representative in a $k$-components model.
A vast array of techniques have been explored, both for \kmeans 
\citep{celebi2013comparative} and \kgmm
\citep{kwedlo2013,blomer2016adaptive,you2023new}.
These seeding strategies recently underwent important developments
consisting of improving the initial $K$ seeds via a {\em re-selection}
mechanisms, local searches and
swaps \citep{lattanzi2019better}, \citep{choo2020k}, \citep{fan2023lsdspp}, \citep{grunau2023nearly}.
%%

%% First, the \kmeans seeding techniques are relevant for \kgmm, since
%% the clusters induced by the seeds can be used
%% to infer the initial Gaussian components.
%% %%
%% Both deterministic and randomized seeding techniques have been
%% developed, but the (greedy) randomized smart seeding
%% of \kmeanspp \citep{arthur2007k} and its recent variants based on
%% multi-swaps
%% \citep{grunau2023nearly} are both simple and effective.

\paragraphmini{Contributions.}  We design novel seeding methods for \kmeans
and \kgmm yielding (i) a lesser number of (Lloyd, EM) iterations, and
(ii) a lesser variability in the output.
To achieve these goals, we revisit the previous seeding methods and
formalize their three key ingredients: the metric used to sample candidate
seeds, the number of seed candidates, and the metric used to rank candidate
seeds.  This analysis brings out two general design principles for
seeding methods.
The first is a {\em lookahead} principle, which consists of
conditioning the seed selection to an enhanced coherence with the
final metric used to assess the algorithm.
The second is a {\em multipass strategy}, which consists of performing
the seed selection in at least two passes with a specific ordering, to tame down the effect of
randomization.

\begin{figure}[htb]% or !htb or H
\centering
\fbox{\includegraphics[width=0.45\textwidth]{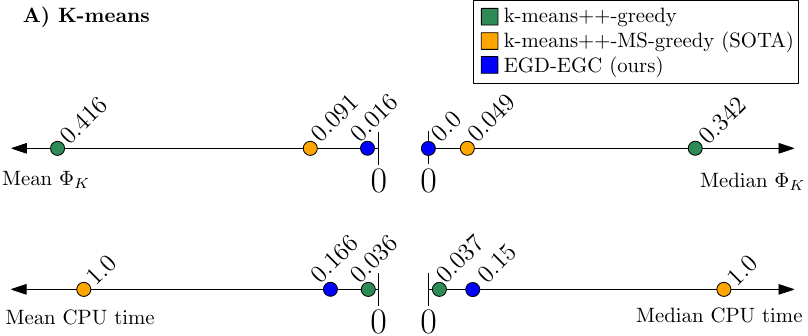}}
\fbox{\includegraphics[width=0.45\textwidth]{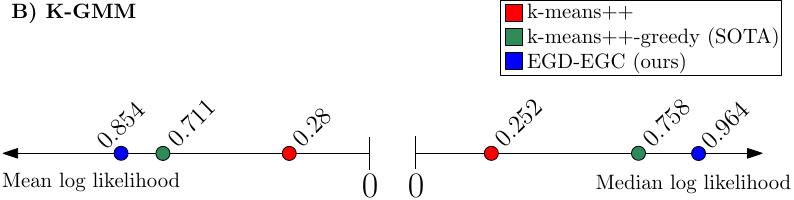}}
\caption{{\bf Gains yielded by our seeding methods.}  {\bf (A,
    \kmeans)} Mean and median (over 18 datasets) of min-maxed SSE
  $\kmeansfunc$ ($\mmM$, see Sec. \ref{sec:stats}) and CPU time ($\mmM{\overline{t}}$,
see Sec. \ref{sec:stats}), the smaller the better--see also
  Sec.~\ref{sec:kmeans-results}.  {\bf (B, \kgmm)} Mean and median of
  min-maxed Log likelihood (over 1800 datasets), the larger the
  better--see also Sec.~\ref{sec:EM-results}. Seeding methods have
  negligible impact on CPU time for \kgmm (data not shown) }
\label{fig:graphical-abstract} 
\end{figure}

Our methods bear three major differences with the recently
developed reselection
schemes~\citep{lattanzi2019better,choo2020k,fan2023lsdspp,grunau2023nearly,huang2024linear}.
The first is the {\em metric} used to perform selection, which is
the distance to the centroids of the clusters induced by the centers--instead
of the distances to the centers themselves. In spirit, this strategy is consistent with
early work on \kmeans, also using centroids to obtain complexity
\citep{inaba1994applications} and approximation bounds~\citep{matouvsek2000approximate}.
%%
%% The second one is the size of the pool from which the re-selection is
%% carried out--as opposed to multi-swap methods we never work with a
%% pool of seeds of size $>K$.
%%
The second one is the reselection strategy, based on the addition
rather than the removal of seeds--we never work with a pool of seeds
of size $>K$.
The last one is the imposed ordering in which we perform the reselection.
Overall, our design choices yield reselection schemes which outperform
the recently developed multi-swap strategy from \citep{grunau2023nearly}, 
 (Fig. \ref{fig:graphical-abstract}), especially when it comes to running times.

\vspace{\cheatheight}

%% short version in there
%%\input{previous-work.tex}
%%\clearpage

\section{Previous work on \kmeans}
%%i%%%%%%%%%%%%%%%%%%%%%%%%%%%%%%%%%%%%%%%%%%%%%%%%%%%%%%%%%%%%%%%%%%%%%%%%%%%%%%%

\subsection{\kmeans}
%%ii-%-%-%-%-%-%-%-%-%-%-%-%-%-%-%-%-%-%-%-%-%-%-%-%-%-%-%-%-%-%-%-%-%-%-%-%-%-%-%

\paragraph{\kmeans and its complexity.}
In \kmeans, let $c_i$ be the center of mass (COM) of the $i$-th
cluster $C_i$.  The Sum of Squared Errors (SSE) functional reads as
\begin{equation}
\label{eq:SSE}
\kmeansfunc = \sum_{i=1,\dots,K} \sum_{x_j\in C_i} \vvnorm{x_j-c_i}^2.
\end{equation}
\ifLONG
From the combinatorial standpoint, assigning $n$ points to $K$
clusters yields $K^n$ possibilities, but this gross enumeration
contains clusterings with empty clusters.  
A sharper analysis shows
that the number of partitions into $K$ nonempty clusters is given by
the so-called Stirling number of second kind,
which, for fixed $K$, behaves as $K^n/k!$.
Getting further insights into the complexity of the problem requires
discussing whether the values of $d$ and $K$ are fixed or not.
(I) Assume $d$ and $K$ are fixed. As noted above, a \kmeans clustering
is implicitly defined by the Voronoi diagram of the $K$ centers in the
$d$-dimensional space.
By  enumerating  all such partitions and computing the 
functional of Eq. (\ref{eq:SSE}), one solves \kmeans exactly.
It turns out that   the maximum number of
possible partitions of $n$ data points generated by the Euclidean Voronoi diagram is
$O(n^{O(dk)})$~\citep{inaba1994applications}, so that \kmeans is
solvable in polynomial time.
(II) Assume now that $n$ and $d$ are free, but $K=2$ is fixed. Then,
constructing instances of 2-means with $n$ points in
dimension $d=2n$, it has been shown that \kmeans is
NP-hard~\citep{dasgupta2008hardness}.  (III) Finally, with $d=2$, assume that $n$
and $K$ are free--that is there is no bound on $K$ which can be
linear in $n$.  Then planar \kmeans is also 
NP-hard~\citep{mahajan2012planar}.
\else
From the geometric standpoint, if one assumes that  $d$ and $K$ are fixed,
\kmeans is solvable in 
 $O(n^{O(dk)})$ polynomial time~\citep{inaba1994applications}.
However, if $k$ or $d$ are functions on $n$, \kmeans
is NP-hard~\citep{dasgupta2008hardness,mahajan2012planar}.
\fi

From a practical standpoint, so-called {\em Lloyd} iterations are used
to improve an initial set of seeds~\citep{lloyd1982least}, by iterating
two steps: (i) ascribe each data point to its nearest center, (ii)
recompute the center of mass of each cluster. The process halts when
the clusters are stable.  The outcome naturally depends on the initial
choice of seeds--it is a random variable--and no information is
provided with respect to the optimal value of $\kmeansfun$, denoted
$\kmeansopt$.

\paragraph{Randomized seeding with  \kmeanspp.}
A landmark has been the design of the $\kmeanspp$ {\em smart} seeding
strategy, which consists of ensuring that the initial centers are
correctly placed in the unknown clusters~\citep{arthur2007k}.
Assume a set of seeds $S_k$ has been selected, and for each sample in
$x\in X\backslash S_k$, let $\Dsquare{x}$ be the square of the minimum
distance to a seed. The next seed $c_{k+1}$ is chosen at random from
$X\backslash S_k$ using the probability $\Dsquare{\cdot}$.
Under this selection scheme, the outcome $\kmeansfun$ is a random variable
satisfying  $\expX{\kmeansfun} / \kmeansopt \leq 8 (\ln K  + 2)$~\citep{arthur2007k}.
A useful heuristic also  described in~\citep{arthur2007k}
consists of choosing each new seed as the best out of a pool of size $l$.
%% \quoteen{Also, experiments showed that \kmeanspp generally
%% performed better if it selected several new centers during
%% each iteration, and then greedily chose the one that
%% decreased $\phi$ as much as possible.}
%%
This seeding variant, referred to as {\em greedy \kmeanspp} or
\kmeansppg \citep{celebi2013comparative}, is implemented in
\sklearn with $l = 2+\log K$ candidates.
The method has approximation factors of $O(l^3\log^3 k)$ and
$\Omega(l^3\log^3 k / \log^2(l\log
k))$ \citep{bhattacharya2020noisy,grunau2023nearly}. It is
theoretically preferable to use a single seed, as increasing the pool
size reduces randomization whence the quality of seeds.
%%a rather   counter-intuitive re

\paragraph{Improved seeding with reselectors.}
So-called {\em local searches} (LS) consist of replacing seeds by
samples when $\kmeansfun$  decreases~\citep{kanungo2002local}.
%%and an exhaustive search of LS yields an optimal constant approximation factor of $9+\eps$.
%%
To replace one seed, the \kmeansppls algorithm samples the new
candidate seed with the $\Dsquare$ strategy instead of checking all
possible options~\citep{lattanzi2019better}.  Running
$Z=\Omega(k\log\log k)$ iterations yields a constant approximation
factor (CFA) of 509 \citep{lattanzi2019better}, later improved to $\sim
26.64$ \citep{grunau2023nearly}. The number of iterations to obtain
such a CFA has also been studied \citep{choo2020k}, as well as the
complexity of the method \citep{fan2023lsdspp}.  In practice, though,
\kmeansppg outperforms these improvements--see our Experiments.
Last but not least, the {\em multi-swap} variant consists of opting
out $p>1$ seeds instead of one~\citep{beretta2023multi}, yielding the
\kmeansppms algorithm. Using $p=O(1)$ and $Z=O(ndk^{p-1})$ iterations yields a CFA $<10.48$.
Practically, exploring the $\binom{k+p}{p}$ candidate swaps is not
 effective.  Starting from $K+p$ seeds, a greedy variant iteratively
 discards the seed minimizing the cost increase--thus less 
 representative of the data.  This greedy version,
 denoted \kmeansppmsg, outperforms \kmeansppg in
 practice~\citep{beretta2023multi}.  But as we shall see, it is
 outperformed by our seeders, especially for the running time.

\paragraph{Deterministic seeding.}
\ifLONG
Beyond \kmeanspp and the recent variants, two deterministic methods
are worth mentioning.  The first one, \kmeanspca, is a recursive
partitioning of the cluster with largest variance along the first
principal direction, until $K$ groups of points have been
obtained~\citep{su2007search}. The initial centers are the COM of these
point sets.
The second one, \kmeansca, replaces the principal direction by the
coordinate axis with the largest variance, a simplification aiming at
avoiding the burden of the covariance matrix calculation or that of
the largest eigenvalue/vector.

\else
Deterministic seeding methods have also been proposed, in
particular \kmeanspca and \kmeansca~\citep{su2007search}.  However,
they require costly operations, and \kmeanspp often performs on par
with them.
\fi

A  thorough experimental comparison has been presented in \citep{celebi2013comparative},
using 32 datasets up to $\sim 2$ M points and dimension up to $d=617$.
Three methods consistently outperform the remaining ones
(Fig. \ref{fig:celebi-table2}): \kmeanspp, \kmeansppg, \kmeanspca.

\section{Previous work on Gaussian Mixture fitting using EM}

Gaussian mixtures models (GMM) are of fundamental interest, both in
theory and in practice.  We briefly review below recent results on the
learnability of GMM and the role of seeding.

\paragraph{Learnability and connexion to seeding.}
A \kgmm model is defined by as a weighted sum of Gaussian
distributions, that is $\normalD{x}{\Theta} = \sum_{k=1}^K w_k
\normalD{x}{\mu_k, \Sigma_k}$, with $\sum_k w_k = 1$.
The learnability of GMM received a considerable attention, and we only mention
the most recent papers we are aware of, in two veins.
The first vein deals with the computation of a GMM close to the
unknown one in total variation (TV) distance.  Assuming a lower bound
on the mixing weights and also on the pairwise TV distance between the
components, a probabilistic polynomial time is
possible~\citep{liu2023robustly}.  It exploits certain algebraic
properties of the higher order moments of the Gaussians.  But as far
as we know, such approaches are not of practical interest.
The second vein is the learnability via the recovery of clustering
labels; that is, assuming that the samples have been generated by a
GMM, one wishes to identify which Gaussian generated which sample.
Optimal clustering rates were recently
reported~\citep{chen2024achieving}, based on separability hypothesis on
the components involving the Mahalanobis distance between the
centers. The algorithm uses a hard EM starting from a {\em decent}
initialization, namely a classifier with sublinear loss.  Such a warm
start is achieved using the vanilla Lloyd algorithm--see
also \citep{gao2022iterative}, and our methods are of direct interest
for this step.

\paragraph{Initialization of EM for GMM fitting.}
When a GMM is fitted using EM, be it soft EM~\citep{bishop2006pattern}
or hard EM~\citep{chen2024achieving}, the outcome depends on the
initial mixture, whose design received a significant
attention--see \citep{kwedlo2013,blomer2016adaptive,you2023new} and
references therein.  In short, the reference methods obtain a
partition of the dataset and use the points of the corresponding
clusters to estimate the mixture components passed to EM.  An
interesting observation is that it is often beneficial to estimate
isotropic initial components instead of anisotropic ones
(Algorithms \meanstosphgmm and
\meanstogmm), Algo. \ref{alg:means2gmm} and \citep{blomer2016adaptive}.
Intuitively, along the greedy selection process, the
parameters estimated at step $k-1$ are only a very coarse estimate of
the $(k-1)$-components of an optimal \kgmm.

The initial clustering can be obtained using \kmeanspp, yielding the
initialization \kgmmseedingPP.  However and as recalled in Introduction,
since a \kgmm is to be estimated, it is interesting to replace
the Euclidean distance by alternative better suited to Gaussian
components.  Of particular interest is the seeding method from
\citep{kwedlo2013}, as the center of a component is iteratively chosen
to maximize the Mahalanobis distance to the already chosen center.
Two generalization were proposed in  \citep{blomer2016adaptive}:

\noindent $\bullet$ {\tt The Spherical Gonzalez (SG)} method chooses a seed by maximizing the Mahalanobis
  distance to the components already chosen. Doing so faces the risk
  of choosing outliers, so that the method samples candidates in
  $S\subset X$, with $\size{S} = \lceil s\size{X} \rceil$ -- with
  $s$ a hyperparameter $\in (0,1]$.

\noindent $\bullet$ {The {\tt Adaptive (Ad)} method chooses points using a strategy
similar to $\Dsquare$, except that the probability distribution used
mixes an $\alpha$ component of the Mahalanobis distance, and a
$1-\alpha$ fraction of the uniform distribution on points.
This latter component aims at avoiding outliers.

A further independent  option has been studied. As noted above, at
each step, \meanstosphgmm is used to define the $k$ mixture
components.  In addition, one classification EM step (CEM) may be
used to refine the mixture~\citep{celeux1992classification}. (CEM can
be seen as a classification version of EM, as it imposes a hard
classification step between the E-step and the M-step of EM.)
%%

%%i%%%%%%%%%%%%%%%%%%%%%%%%%%%%%%%%%%%%%%%%%%%%%%%%%%%%%%%%%%%%%%%%%%%%%%%%%%%%%%%
%%i%%%%%%%%%%%%%%%%%%%%%%%%%%%%%%%%%%%%%%%%%%%%%%%%%%%%%%%%%%%%%%%%%%%%%%%%%%%%%%%

\section{New seeding strategies for \kmeans}
\label{sec:new-seeding-kmeans}
%%i%%%%%%%%%%%%%%%%%%%%%%%%%%%%%%%%%%%%%%%%%%%%%%%%%%%%%%%%%%%%%%%%%%%%%%%%%%%%%%%

\subsection{Selected useful observations}
%%ii-%-%-%-%-%-%-%-%-%-%-%-%-%-%-%-%-%-%-%-%-%-%-%-%-%-%-%-%-%-%-%-%-%-%-%-%-%-%-%

\paragraphmini{Notations for Sum of Squared Errors.} The following notations will be useful:

\sbullet{$\kmeansfun$:} the \kmeans SSE functional upon termination of \kmeans -- Eq. \eqref{eq:SSE}.

\sbullet{$\kmeansfunS$:} the SSE of data points, using for each such point the
  distance to its nearest seed--which is also a data point. That is,
  Eq. \eqref{eq:SSE}, where the centers are the $K$ seeds.

\sbullet{$\kmeansfunCOM$:} the SSE of data points, using for each such point
  the distance to the center of mass (COM) of all samples sharing the same seed.  That
  is, Eq. \eqref{eq:SSE}, where the centers are the COM of the $K$ clusters
associated with $K$ seeds selected.

\paragraphmini{$\Dsquare$ distances during seed selection.}
Another interesting parameter is the stability
of the distribution of distances used by the $\Dsquare$ strategy.
The typical behavior is a decreasing variance of the mean (over all
points) squared distance $\Dsquaremean$ along the seed selection
(Fig. \ref{fig:d2-boxplot}). This indicates that distance-wise, the
choice of seeds with a large index $k\in 1,\dots,K$ is more stable
 along seed selections  than that of seeds with
low index.

\paragraphmini{$\kmeansfunS$ and $\kmeansfunCOM$ during seed selection.}
We replicate the analysis just carried out for $\Dsquaremean$ to
$\kmeansfunS$ and $\kmeansfunCOM$
(Fig.~\ref{fig:greedy-metrics-boxplot} and
Fig.~\ref{fig:greedy-metrics-scatter}). $\kmeansfunS$ has a behavior
similar to that of $\Dsquaremean$. This suggests that \kmeansppg
suffers from the limitation seen with \kmeanspp, as the metric used to
compare candidate seeds also stabilizes when the index $k$
increases. Interestingly, statistics for $\kmeansfunCOM$ are much more
stable  along seed selections, with fewer outliers and more concentrated boxes.

%\ifLONG
\begin{figure*}%[htbp]
\begin{center}
\begin{tabular}{cc}
\includegraphics[width=.5\textwidth]{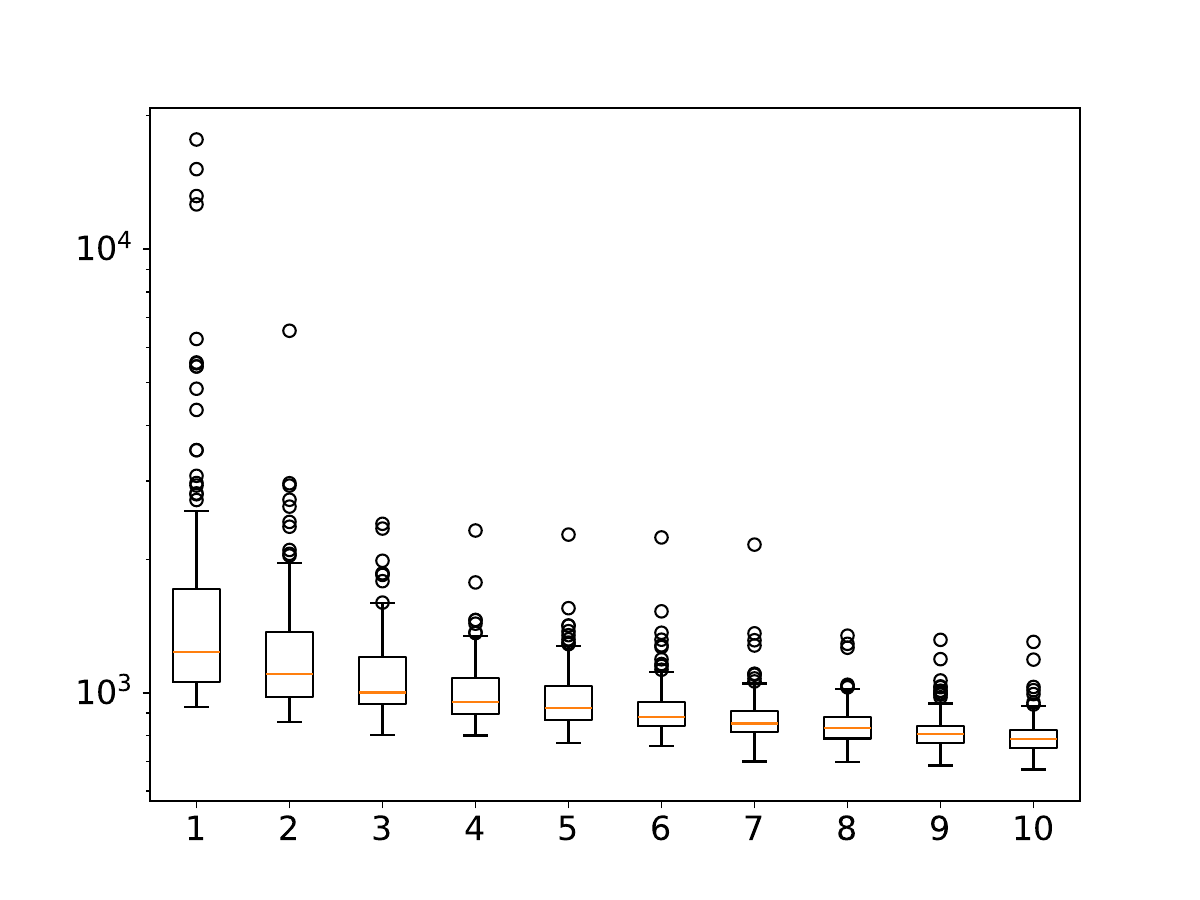}&
\includegraphics[width=.5\textwidth]{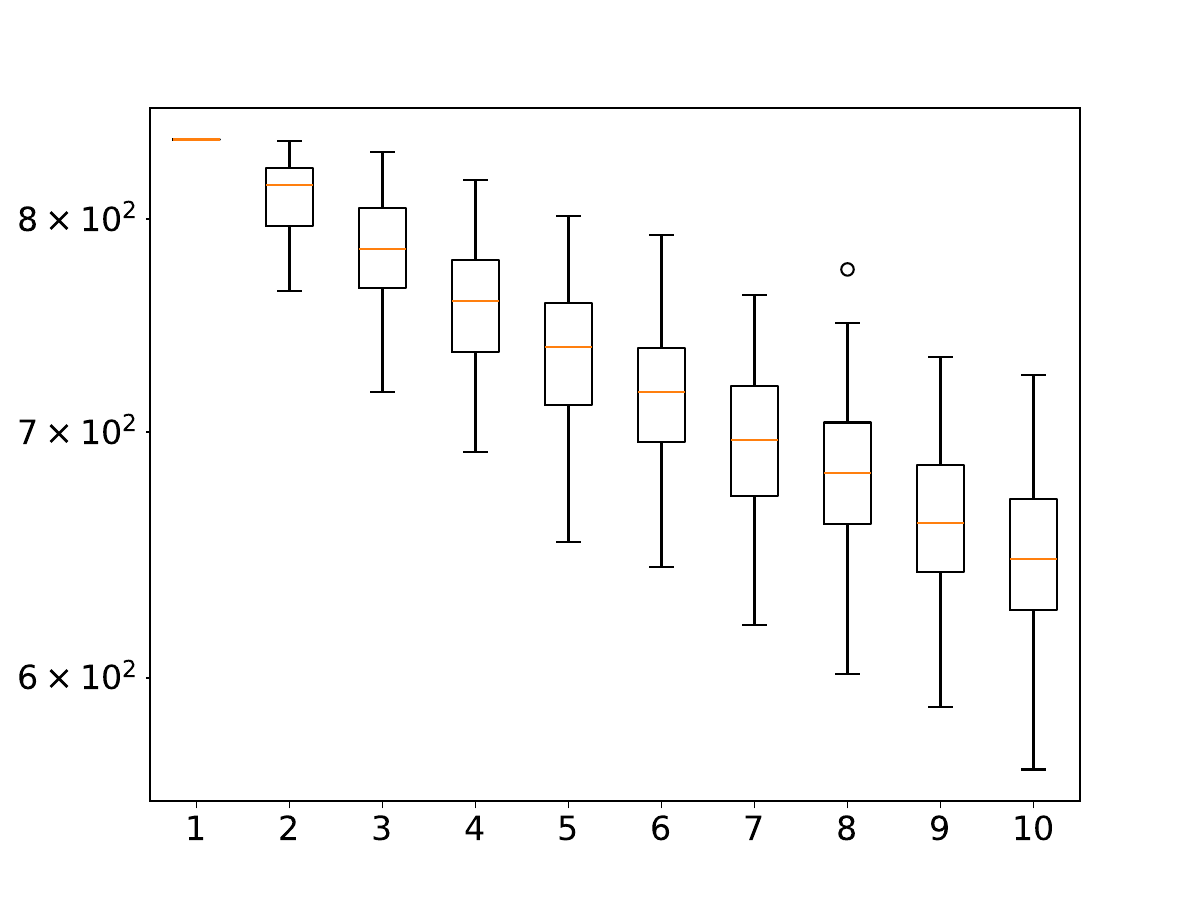}\\
$\kmeansfunS$ & $\kmeansfunCOM$\\
\end{tabular}
\end{center}
\vspace{-0.5cm}
\caption{{\bf \kmeans: boxplot of the $\kmeansfunS$ and $\kmeansfunCOM$ values along the seeding selection process
for each $k\in 1,\dots,K$.}
Statistics over 150 repeats on \textit{spam} dataset.}
 \label{fig:greedy-metrics-boxplot}
\end{figure*}

%% \else
%% \vspace{\cheatheight}
%% \begin{figure}[htbp]
%% \begin{center}
%% \begin{tabular}{cc}
%% \rotatebox{90}{\quad \quad \quad \quad \quad $\kmeansfunS$}&
%% \includegraphics[width=.35\textwidth]{fig/phikd_boxplot}\\
%% %%
%% \rotatebox{90}{\quad \quad \quad \quad \quad $\kmeansfunCOM$} &
%% \includegraphics[width=.35\textwidth]{fig/phikscom_boxplot}
%% \end{tabular}
%% \end{center}
%% \vspace{-0.5cm}
%% \caption{{\bf \kmeans: boxplot of the $\kmeansfunS$ and $\kmeansfunCOM$ values along the seeding selection process
%% for each $k\in 1,\dots,K$.}
%% Statistics over 150 repeats on \textit{spam} dataset.}
%%  \label{fig:greedy-metrics-boxplot}
%% \end{figure}
%% \fi

\subsection{Notations for seeding variants}
%%ii-%-%-%-%-%-%-%-%-%-%-%-%-%-%-%-%-%-%-%-%-%-%-%-%-%-%-%-%-%-%-%-%-%-%-%-%-%-%-%

The seeding in \kmeanspp and \kmeansppg selects $K$ seeds in one pass
using distances from data points to already selected seeds.  We
propose a multipass seed selection strategy, each pass being qualified
by three  ingredients:

\sbulem{(Options: sampling candidate seeds)} E: Euclidean distance.  (NB: 
used for the sake of coherence with seeding methods
  for \kgmm, see Section \ref{sec:seeding-EM}.)

\sbulem{(Options: size of the pool of candidate seeds)} O: One | G: Greedy.  The
  symbol O (resp. G) refers to a seed selection using a single
  (resp. $\log K+2$) candidates--this latter number being that used in
  the \sklearn implementation.

\sbulem{(Options: ranking candidate seeds)} D: Data | C: COM | N:
  NA.  The letter D (resp. C) refers to a selection using the distance
  between data points (resp. from data points to the centers of masses
  induced by the seeds). For a pool of size one, there is no such
  design strategy -- whence N or Not Applicable.

Summarizing, a seed selection process is described by the following
regular expression:
\ifTWOCOLUMNS
\begin{equation}
\scriptsize \seeding{\{E\ \ [\text{O|G}]\ \ [\text{D|C|N}] \}}^+.
\end{equation}
\else
\begin{equation}
\seeding{\{E\ \ [\text{O|G}]\ \ [\text{D|C|N}] \}}^+.
\end{equation}
\fi
Let us illustrate these conventions with \kmeanspp and \kmeansppg:

\sbulem{\seedingkmeanspp:} the seeding in \kmeanspp.  Seeds are selected in
 one pass; the selection of a seed uses a single candidate (letter O, which 
implies letter N).

\sbulem{\seedingkmeansppg:} seeding used in \kmeansppg.  Seeds selected
  in one pass; the seed is selected amidst a pool of candidates
  (letter G); in this pool, the seed retained is that yielding the
  lowest $\kmeansfunS$~\citep{arthur2007k}, which uses the distance
  from samples to seeds, which are themselves samples (whence the letter D).

\subsection{New iterative seeding strategies}
%%ii-%-%-%-%-%-%-%-%-%-%-%-%-%-%-%-%-%-%-%-%-%-%-%-%-%-%-%-%-%-%-%-%-%-%-%-%-%-%-%

\paragraphmini{Variance reduction: \seedingEGDEGD.}
\kmeanspp uses a single pass strategy.  To mitigate the influence
of the initial steps, we propose a two-pass
zig-zag selection process of seeds (Algorithm \ref{alg:SeedingEONEON}).
We have seen that the variance of $\Dsquare$ distances 
along seed selections  decreases during the seed selection
(Fig. \ref{fig:d2-boxplot}).
Therefore, by reselecting the centers a second time, the $\Dsquare$
selection is less dependent on randomness and thus more accurate.

This second selection can be done in two ways, by processing the seeds
upstream (zig pass, from 1 to $K$), or downstream (zag pass, from $K$
to 1).  Experiments have shown that the latter performs better--data
not shown.
Combining this multi-pass strategy with a greedy selection results in the \seedingEGDEGD method : the second EGD
zag pass reselects seeds, choosing the best amidst a pool of 
$\log K +3$ candidates ($\log K+2$ candidates as in
\kmeansppg, and the center obtained during the zig pass).

\paragraphmini{Look-ahead: \seedingEGDEGC.}   In anticipation
for LLoyd iterations, we propose to use $\kmeansfunCOM$ instead of
$\kmeansfunS$ to rank the seeds in a pool of candidates.  Indeed, the
first Lloyd iteration replaces the initial seeds by the center of mass
of that cluster.

In practice, this strategy is only effective for large values of $k\in
1,\dots,K$, as $\kmeansfunCOM$ does not sufficiently discriminate
candidates on the first seeds.  The extreme case is that of the first
seed selected, for which the center of mass is unique.  Experiments
confirmed this behavior (data not shown), so that we stick to the seed
variant \seedingEGDEGC, in which we replace $\kmeansfunS$ with
$\kmeansfunCOM$ during the zag pass.

\subsection{Seeding and final objective: correlation?}
\label{sec:seeding-phiK-corr}
%%ii-%-%-%-%-%-%-%-%-%-%-%-%-%-%-%-%-%-%-%-%-%-%-%-%-%-%-%-%-%-%-%-%-%-%-%-%-%-%-%

Given that we aim at optimizing the SSE functional $\kmeansfunc$, the look
ahead principle just outlined seems rational.
However, it is also instrumental to think about the \kmeans problem in
terms of energy / fitness landscape~\citep{dicks2022elucidating}.  To
do so, define the \emph{fitness landscape} of a \kmeans problem as set
of pairs $\{(\Pi_i, \kmeansfunc[i])\}_{i\geq 1}$ obtained during the
Lloyd iterations, with $\Pi_i$ the partition / clustering of the point
cloud, and $\kmeansfunc[i]$ the corresponding SSE. (NB: index $0$
corresponds to the seeding outcome.)
Let us, intuitively, define a {\em funneled} fitness landscape as a
\kmeans problem such that the sequence of partitions $\Pi_i$ visited
during Lloyd iterations eventually leads to a {\em low} lying local
minimum $\kmeansfunc[\text{final}]$, that is a clustering whose SSE is close to the optimal value
$\kmeansopt$.
In that case, different initialization with drastically different
$\kmeansfunc[0]$ may yield to the same value
$\kmeansfunc[\text{final}]$.  Which means that no correlation will be
observed between $\kmeansfunc[0]$ and $\kmeansfunc[\text{final}]$.

To substantiate this intuition, we study the (Pearson and Spearman)
correlations $(\kmeansfun, \kmeansfunS)$ and ($\kmeansfun,
\kmeansfunCOM)$ on classical  datasets  \citep{celebi2013comparative},
using \kmeansppg for $\kmeansfunS$ and \seedingEGDEGC for $\kmeansfunCOM$~(Fig. \ref{fig:correlations}).
A mild correlation is observed is all cases, confirming our expectations.
Importantly, this fact does not contradict the approximation factors discussed in previous
work: the approximation factors qualify the distance to the optimal SSE $\kmeansfunc$,
while the aforementioned correlations depend on the topography of the
\kmeans fitness landscape, or, phrased differently, in the sequence
of partitions $\Pi_i$ visited during Lloyd iterations.

%%
%% First, we observe a mild  correlation between $\kmeansfun$ 
%% and $\kmeansfunS$, a rather striking fact given the 
%% notable role played by $\kmeansfunS$ in \kmeansppg.
%% %%
%% Second, we note that the correlation ($\kmeansfun, \kmeansfunCOM)$ is
%% slightly better than that for $(\kmeansfun, \kmeansfunS)$ on most
%% datasets. This observation underlies our novel seeding strategies,
%% which exploit distances to COM induced by seeds, rather than distances
%% to seeds themselves.

\section{New seeding strategies for \Kgmm}

\label{sec:seeding-EM}
%%i%%%%%%%%%%%%%%%%%%%%%%%%%%%%%%%%%%%%%%%%%%%%%%%%%%%%%%%%%%%%%%%%%%%%%%%%%%%%%%%

\subsection{Notations for seeding variants}

The seeding methods developed for \kgmm follow the multipass strategy
introduced for \kmeans . Yet, the use of a mixture model allows for
new metrics both in the sampling and selection of candidate seeds.
Thus, our new seeding strategies consists in combinations of \kgmm
seeding passes, followed by the use of \meanstogmm to transform
the selected seeds into an initial model to be optimized. (NB: the
seeds are used as $\mu$s arguments in the Algorithm \ref{alg:means2gmm}.)
The \kgmm passes are qualified with three ingredients:

\sbulem{(Options: sampling candidate seeds)} E: Euclidean distance
  | A: Adaptive Mahalanobis | G: Gaussian distance. The symbol E corresponds
  to the Euclidean distance originally used for $\Dsquare$ weighting
  in \kmeanspp.  The symbols A refers to the distances used in
  the Adaptive seeding method (with $\alpha = 0.5$)~\citep{blomer2016adaptive}.
Finally,  G refers to a strategy using the $\Dsquare$ method on distances
(Eq. \ref{eq:dist-gauss}) between Gaussian distributions estimated at every data point.
See details in Section \ref{sec:kgmmseedingGGD}.

\sbulem{(Options: size of the pool of candidate seeds)} O: One |  G:Greedy.  Similar to \kmeans.

\sbulem{(Options: ranking candidate seeds)} D: Data | C: COM | L: Log-likelihood |
  N: NA.  The letters D, C and N match the options used
  for \kmeans. We add a new metric with the letter L, corresponding to
  the log-likelihoods of mixture models estimated using each candidate
  seed.

Summarizing, a seed selection process is described by the following
regular expression:
\ifTWOCOLUMNS
\begin{equation}
\scriptsize \kgmmseeding{\{[E|A|G]\ \ [\text{O|G}]\ \ [\text{D|C|L|N}] \}}^+.
\end{equation}
\else
\begin{equation}
\kgmmseeding{\{[E|A|G]\ \ [\text{O|G}]\ \ [\text{D|C|L|N}] \}}^+.
\end{equation}
\fi

\subsection{New iterative seeding strategies}
%%ii-%-%-%-%-%-%-%-%-%-%-%-%-%-%-%-%-%-%-%-%-%-%-%-%-%-%-%-%-%-%-%-%-%-%-%-%-%-%-%

\subsubsection{Greedy adaptive selection: \kgmmseedingAGL}

The adaptive selection strategy combines Mahalanobis and uniform
distances to select seeds~\citep{blomer2016adaptive}, before running
\meanstosphgmm to obtain the GMM passed to EM. The seeding
\kgmmseedingAGL adds to this strategy a selection based on the
likelihood, a look-ahead with respect to the EM steps.

\subsubsection{EM look-ahead : \kgmmseedingEGDEGL}

In a look-ahead spirit similar to that introduced with $\kmeansfunCOM$
for \kmeans, we use the log-likelihood as a selection metric to rank seeds
among candidates.
To obtain a log-likelihood value from a set of seeds, the \meanstogmm
algorithm is used to construct a temporary Gaussian mixture model~\citep{blomer2016adaptive}. 
Similarly to the use of $\kmeansfunCOM$ for \kmeans, the
log-likelihood is irrelevant to discriminate  candidates for small
values of $k$.  
Consequently, the \kgmmseedingEGDEGL seeding variant uses it for the zag pass only.

\begin{remark}
The observations raised for \kmeans (Sec. \ref{sec:seeding-phiK-corr})
are also valid for \kgmm.
\end{remark}
%% \gc{deprecated, what should we do}
%% We measure the correlation between the log-likelihoods of the
%% initial and final models (\ie before and after the EM iterations), and
%% compare it to the correlation between the $\kmeansfunCOM$ obtained
%% after seeding and the log-likelihood of the final
%% model~(Fig.\ref{fig:correlations-EM}).  Log-likelihood appears as a
%% good selection metric on noiseless data, but suffers from the inclusion
%% of noise.

\subsubsection{EM look ahead and adaptive selection in the zag pass: \kgmmseedingEGDAGL}

We  combine the best features used in the previous two
methods.  First, the adaptive selection using a GMM, as in
\kgmmseedingAGL. The adaptive selection is restricted to the second pass
though, to ensure that the GMM  is representative of the cluster
structure of the data.  Second, the two pass and look-ahead strategy
of \kgmmseedingEGDEGL.
The resulting method is called  \kgmmseedingEGDAGL.

\subsubsection{$D^2_G$ with Gaussian distance}

Finally, we explore the use of \kmeansppg based on a distance between
Gaussians locally estimated at each sample, see method \kgmmseedingGGD
in Section \ref{sec:kgmmseedingGGD}.

\begin{comment}
Adaptive selection relies on the estimation of a GMM using the already
selected seeds. This is done through the \meanstogmm
algorithm. However, this algorithm depends on the induced clustering
of the seeds, which is unrealistic when obtained with the first few
seeds only. For example, the first seed, corresponding to the mean of
the first estimated GMM, will always be the center of mass. Although
this seed is moved overtime as each iteration updates the GMM, it may
still introduces a bias in the seeding. \kgmmseedingEGDAGL restricts the

use of adaptive selection to an unbiased second pass exclusively, by
adding a first pass with traditional \kmeanspp.
\end{comment}

\ifLONG
\section{\kmeans seeding: implementation and tests}
\else
\section{\kmeans seeding: tests}
\fi
\label{sec:kmeans-results}
%%i%%%%%%%%%%%%%%%%%%%%%%%%%%%%%%%%%%%%%%%%%%%%%%%%%%%%%%%%%%%%%%%%%%%%%%%%%%%%%%%

\subsection{Experimental protocol}
\label{ssec:kmeans-exp-protocol}
%%ii-%-%-%-%-%-%-%-%-%-%-%-%-%-%-%-%-%-%-%-%-%-%-%-%-%-%-%-%-%-%-%-%-%-%-%-%-%-%-%

\paragraphmini{Implementation and stop criterion.}  We compare our
seeding methods with \kmeansppls (with $Z=k$) and \kmeansppmsg
(with $Z=k$, $p=2+log(k)$) to initialize \kmeans. We chose
$Z=k$ to match the number of swaps performed by our methods, and
$p=2+log(k)$ to match the candidate pool size originally proposed for
greedy kmeans++ in \citep{arthur2007k}.  All methods were implemented
in C++ using the Eigen library. The Lloyd iterations are stopped when
the Frobenius norm of the difference in the center clusters is smaller
than $1e-4$. As a failsafe, the Lloyd iterations are also stopped
after reaching a maximum number of iterations of 50.
A proof-of-concept python implementation of the best
contender, \seedingEGDEGC, is available from \url{https://anonymous.4open.science/r/multipass-seeding-F03E}. Full implementation to be released upon publication of the
paper.

\paragraphmini{Datasets.}
Our experiments involve 12 datasets from the UCI Machine Learning 
Repository, 11 of which are a subset of the 32 used in 
~\citep{celebi2013comparative}: \{{\itshape Cloud cover, Corel 
image features, Steel plates faults, 
Letter recognition, Multiple features, Musk (Clean2), Optical digits, 
Pen digits, Image segmentation, Shuttle (Statlog), Spambase, Yeast}\}
\footnote{\textit{Spambase} is the one dataset not used in
~\citep{celebi2013comparative}}.

These 12 datasets are the most challenging ones, due to variability
incurred by the seeding strategies, and its effect
on the final clustering~\citep{celebi2013comparative}. The value of $K$ 
for a dataset is that provided alongside each dataset.
They range in size from 1024 to 58,000 data points.

We also process the \{KDD-BIO, KDD-PHY, RNA\} datasets
from \citep{lattanzi2019better} used in the assessment of the SOTA
method \kmeansppmsg \citep{beretta2023multi}. They range in size from
100k to 485k points.  Following \citep{lattanzi2019better}, we cluster
them with $K = \{25,50\}$.

%% In addition, our experiments involve the three datasets originally used in
%% \citep{lattanzi2019better} : \{KDD-BIO, KDD-PHY, RNA\}. 
%% These datasets are included to provide comparison with the current
%% SOTA method \kmeansppmsg, as they are used by the authors to provide
%% experimental results \citep{beretta2023multi}. They are larger than the
%% datasets from~\citep{celebi2013comparative}, ranging in size from
%% 100,000 to 484,565 data points. We follow the experimental protocol
%% of \citep{lattanzi2019better}, such that the values of $K$ used for
%% these datasets are $K = \{25,50\}$.

In total, we investigate $12 + 3 + 3 = 18$ datasets. Following common
practice, on a per dataset basis, we perform a min-max normalization
on the coordinates to avoid overly large ranges. 

\paragraphmini{Hardware.} 
Calculations were run
an a HP desktop computer running Fedora Core 39,
equipped with 24 CPUs (i9-13900K) and 131 GB or RAM.

\subsection{Statistical assessment}
\label{sec:stats}

\paragraphmini{Chosen statistics.}
To compare the various methods, we consider the SSE
$\kmeansfun$ upon convergence of the Lloyd iterations. 
We also measure the CPU time of each seeding strategy, as well as the
average number of Lloyd iterations needed for convergence in the
following \kmeans.  As the seeding methods are non-deterministic, 
we report average results over a set of $R=100$ repeats.

\paragraphmini{Comparison across datasets and normalization
  issues.}  To compensate the variability of statistics across
datasets, we min-max normalize the statistics of interest
($\kmeansfunc$ and running time $t$) in two ways.  Let $m\in \calM$
the particular method to be assessed  and $\calR$ the set of repeats
of that method on a dataset.
To compare methods, the first normalization uses the average values
(computed over repeats), assigning the value 0 (resp. 1) to the worst
(resp. best) method on a dataset -- with $\mmM{\cdot}$ standing for
min-max-Mean:
\begin{equation}
\label{eq:min-max-mean}
\mmM{ \kmeansfuncmean{K, m} } 
= \frac{ \kmeansfuncmean{K, m} -\min_{m'\in \calM} \kmeansfuncmean{K, m'}} 
       { \max_{m'\in\calM} \kmeansfuncmean{K, m'} -\min_{m'\in\calM} \kmeansfuncmean{K, m'}}
\end{equation}
To compare the methods on the range of possible values observed for a given
dataset, the second normalization reads as -- with $\mmG{\cdot}$ standing for min-max-Global:
\begin{equation}
\label{eq:min-max-glob}
\mmG{ \kmeansfuncmean{K, m} } 
= \frac{ \kmeansfuncmean{K, m} -\min_{m'\in \calM, r\in \calR} \kmeansfunc{K, m', r}} 
       { \max_{m'\in\calM, r\in\calR} \kmeansfunc{K, m', r} -\min_{m'\in\calM, r\in\calR} \kmeansfunc{K, m', r}}
\end{equation}
For the running times, we  define likewise $\mmM{\bar{t}_m}$ and  $\mmG{\bar{t}_m}$.

\subsection{Results}
%%ii-%-%-%-%-%-%-%-%-%-%-%-%-%-%-%-%-%-%-%-%-%-%-%-%-%-%-%-%-%-%-%-%-%-%-%-%-%-%-%

\paragraphmini{Incidence of the seeding on $\kmeansfun$.}
We study the contenders from Section \ref{sec:new-seeding-kmeans}.  To
assess the changes brought by the zig-zag strategy itself, we
involve \seedingEGDtwo, namely
\kmeansppg where we double the amount of candidates.
%%
%% We compare the final SSE $\kmeansfun$ yielded by the various methods.
%% Since the various datasets have drastically different values, all SSE
%% are min-max normalized. On a per dataset
%% basis, min-max normalization means that the best (resp. worst) method
%% gets 0 (resp. 1).

Three observations stand out (Fig. \ref{fig:graphical-abstract},
Fig.~\ref{fig:kmeansfunc-means}, Fig.~\ref{fig:kmeansfunc-means2},
SI Table~\ref{tab:results_table_kmeans} (exact values)).

%% \begin{comment}
%% That is, we consider the value $x/max$, with $x$
%% the value for a particular method and $max$ the maximum value on the
%% considered dataset, yielding a maximum value of one.
%% \end{comment}

\vspace{\cheatheight}
\begin{figure}[htbp]
\centerline{ \includegraphics[width=.8\linewidth]{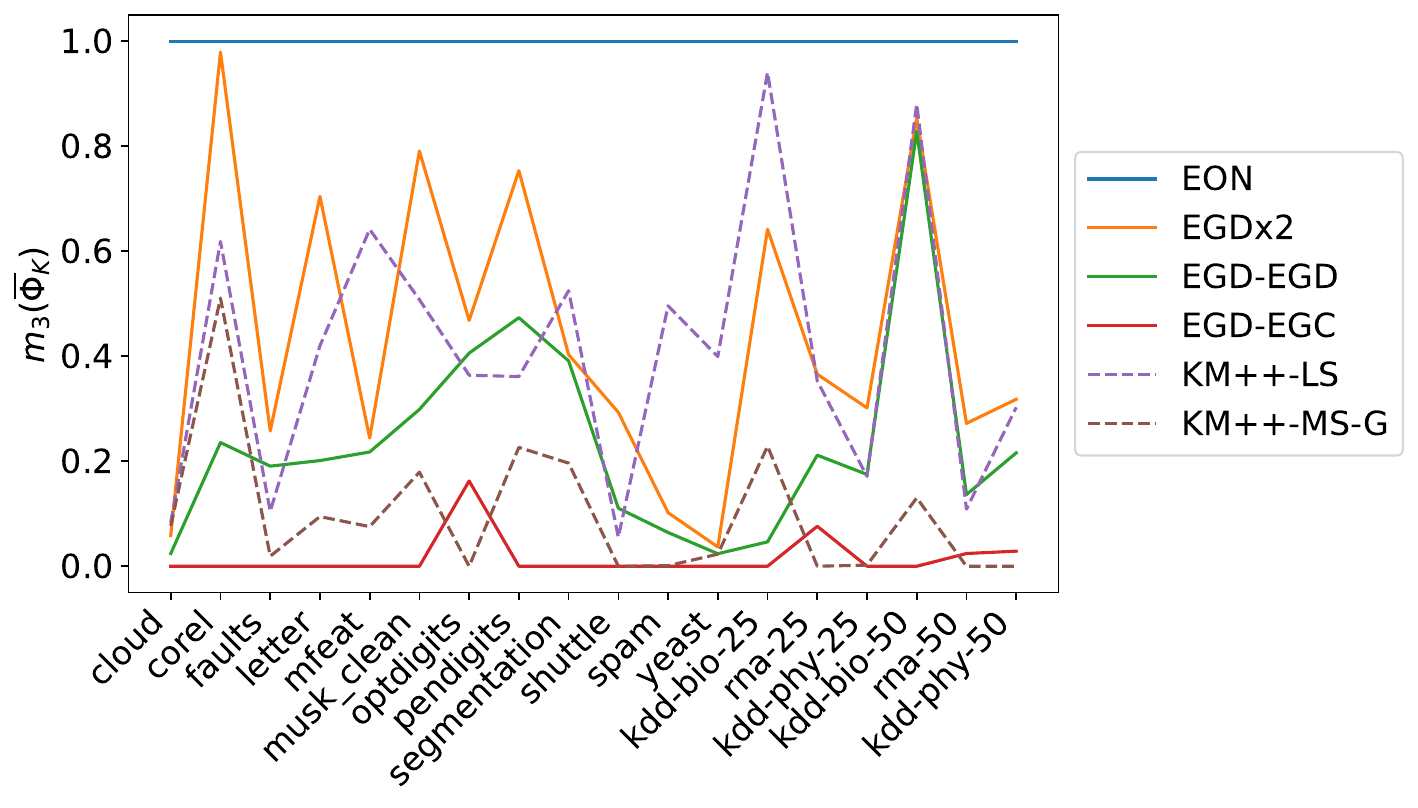}}
\caption{{\bf \kmeans: min-max normalized value $\mmM$ -- Eq. \eqref{eq:min-max-mean},
as a function of the seeding method.} 
For the reference:  the seeding used in \kmeansppg is \seedingkmeansppg,
and \seedingEGDtwo is the same with twice as many seeds to match the zig-zag
strategy.}
\label{fig:kmeansfunc-means}
\end{figure}

\smallskip
\sbulem{Best method.}
In terms of $\mmM$, \seedingEGDEGC outperforms all contenders
 on most datasets, including \kmeansppmsg
 (Fig.~\ref{fig:kmeansfunc-means}, SI Table
 ~\ref{tab:results_table_kmeans}). The improvement is even more
 pronounced in terms of running time
 (Fig.~\ref{fig:seeding-time-kmeans},
 Fig.~\ref{fig:num-Lloyd-iterations-kmeans}).

\begin{comment}
For an overall quantitative assessment, let $\delta$ be the decrease
of $\kmeansfunc$ between \kmeansppmsg and \seedingEGDEGC, and likewise
$\Delta$ the difference between \seedingkmeanspp (\kmeanspp)
and \kmeansppmsg The min, median, mean (stdev), and max values of
$\delta /\Delta$ over our datasets are -0.16, 0.02, 0.12 (0.24),
1.04. The same ratios on the running times result in min,
median, mean (stdev), and max values of $\delta/\Delta$ of: 0.99,
6.30, 6.27 (2.72), 12.37.  (Fig. \ref{fig:graphical-abstract}).
\end{comment}

\sbulem{The zag pass is beneficial when using greedy
  strategies.}  The $\Dsquare$ weighting coupled to the greedy seed selection
  significantly benefits from the zag pass.  As shown with the consistent improvements on several datasets observed with 
  \seedingEGDEGD and \seedingEGDEGC . Specifically, the zig-zag methods
  outperform \seedingEGDtwo on most datasets, providing better
  seeds while considering the same amount of candidates.

\sbulem{Using center of masses to select seeds yields superior
  results.}  The method \seedingEGDEGC outperforms \seedingEGDEGD, confirming
  that $\kmeansfunCOM$ is a better fit than $\kmeansfun$.

\paragraphmini{Running time.}
We study in tandem the min-max normalized CPU total times
(Fig. \ref{fig:graphical-abstract},
Fig.~\ref{fig:seeding-time-kmeans}) and the number of Lloyd
iterations to reach convergence
(Fig.~\ref{fig:num-Lloyd-iterations-kmeans}).
  
\sbulem{Zig-zag seeding methods are slower than single pass equivalents by a factor of 2-3 but
  reduce the number of Lloyd iterations.}
  This trade-off (more
  expensive seeding, less iterations) brings an increase of the total
  running time by x1.5 on the processed datasets, but remains largely
  faster than the current SOTA \kmeansppmsg.
This owes to the following asymmetry: our reselection requires
finding the best seed to add (from a pool of $l$ candidates) once one
seed has been removed; the greedy multi-swap requires finding the best seed
to remove after adding $p$ of them--from a pool of size $K+p$ (first
seed removed) to $K+1$ (last seed removed). 
Moreover, without any optimization / additional storage, the
$\kmeansfunc$ cost update is $O(n)$ for an addition, and $O(Kn)$ for a
removal. 

\sbulem{Seeding methods producing lower values of $\kmeansfun$ also
  tend to reduce the number of required Lloyd iterations.}
We observe an average Pearson correlation coefficient of $0.816$
between $\kmeansfun$ and the number of Lloyd iterations (Fig.~\ref{fig:num-Lloyd-iterations-kmeans}).
This is expected, as efficient seeding precisely aims at placing 
seeds near the optimal positions, reducing the number of Lloyd 
iterations needed for to reach these positions.

\vspace{\cheatheight}
\begin{figure}[htbp]
\centerline{ \includegraphics[width=.8\linewidth]{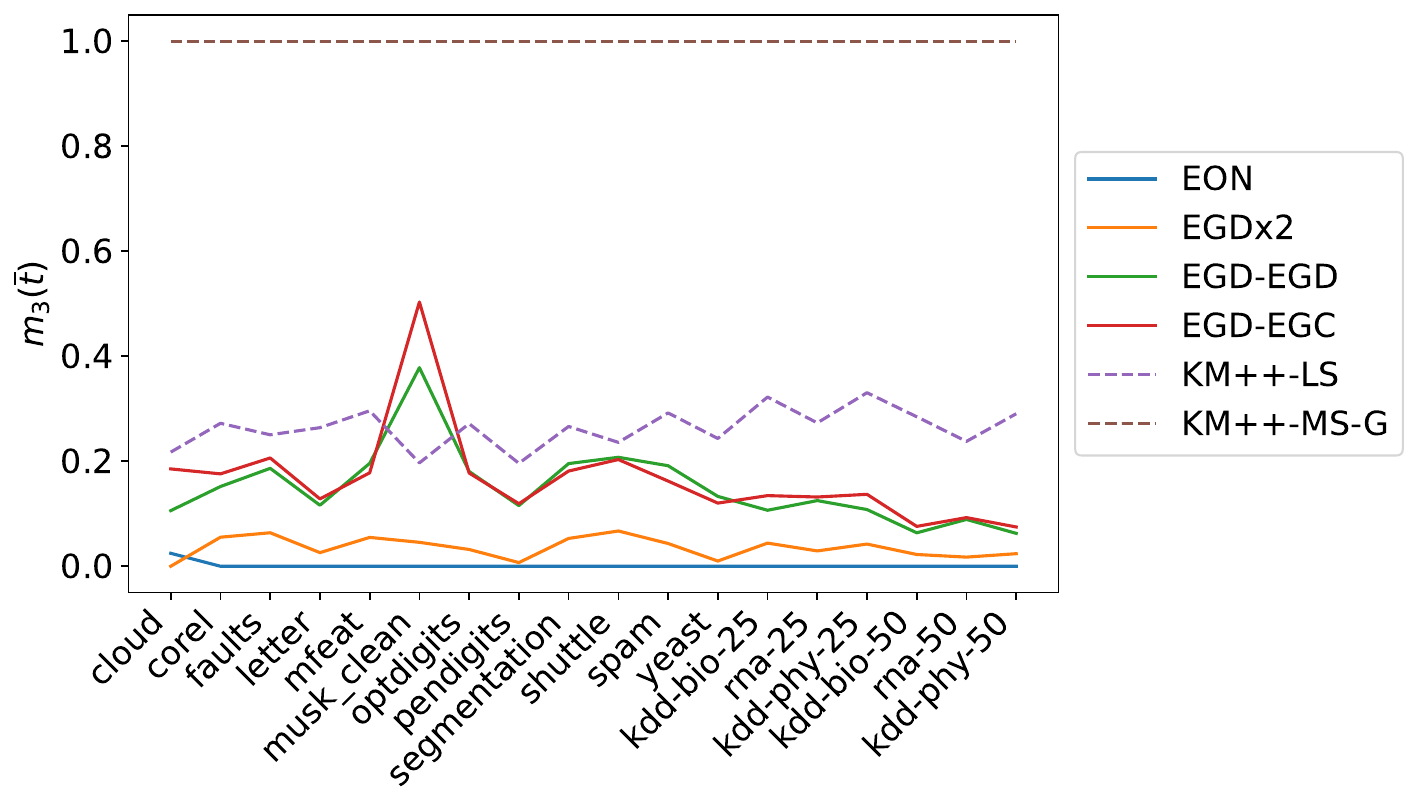}}
\caption{{\bf \kmeans: min-max normalized CPU total time $\mmM{\bar{t}}$ for each seeding method.}} 
\label{fig:seeding-time-kmeans}
\end{figure}

\subsection{Sensitivity to the pool size}

While our experiments were conducted with $\log K+2$ candidates--as in
the standard the \sklearn implementation, it naturally makes sense to
study the sensitivity of the methods to the pool size $l$.  To this
end, we consider three pool sizes on different scales, namely $l\in
\{2+log K, 2+\sqrt{K}, \max(2, K)\}$,  reporting the
min-max normalized values ($\mmM$, Fig.~\ref{fig:parameter_kmeans_func};
$\mmG$, Fig.~\ref{fig:parameter_kmeans_func2}) and CPU times
($\mmM{\bar{t}}$, Fig.~\ref{fig:parameter_seeding_time_kmeans};
raw times, Fig.~\ref{fig:parameter_seeding_time_kmeans2}).

This comparison calls for two comments.
First, the larger the pool size the better the result (and the
longer the running time).  This is somewhat expected for selected
datasets, keeping in mind however that for others, increasing the pool
size tames down randomization, yielding a worse approximation
factor~\citep{bhattacharya2020noisy,grunau2023nearly}.

Second, the span observed for $\kmeansfunc$ decreases when moving from
\kmeansppg to \seedingEGDEGC.  The lesser sensitivity of our method to
the pool size illustrates its ability to identify meaningful seeds at
the onset, a pattern shared by \kmeansppmsg.

\ifLONG
\section{\kgmm seeding: implementation and tests}
\else
\section{\kgmm seeding: tests}
\fi
\label{sec:EM-results}
%%i%%%%%%%%%%%%%%%%%%%%%%%%%%%%%%%%%%%%%%%%%%%%%%%%%%%%%%%%%%%%%%%%%%%%%%%%%%%%%%%

\subsection{Experimental protocol}
\label{ssec:em-exp-protocol}
%%ii-%-%-%-%-%-%-%-%-%-%-%-%-%-%-%-%-%-%-%-%-%-%-%-%-%-%-%-%-%-%-%-%-%-%-%-%-%-%-%

\paragraphmini{Implementation and stop criterion.}
All seeding methods as well as the EM iterations were implemented 
in C++ using the Eigen library. 
\ifLONG
\else
Implementation to be released upon publication of the paper.
\fi

The stopping criterion for the EM iterations targets the relative
difference in log-likelihood between two iterations, and reads as
$\nicefrac{|l(\theta_n) - l(\theta_{n-1})|}{|l(\theta_{n-1})|} <
1e-4$. Alternatively, the EM iterations are stopped after reaching a
maximum number of 100 iterations.

\paragraphmini{Generated datasets.}
Following \citep{blomer2013simple}, we use GMMs  defined
from the following parameters: (i) the separation between components
(values $s=0.5, 1, 2)$, (ii) the weights of components (uniform,
different), (iii) the size of components (constant, different), and
(iv) their eccentricity ($e = \nicefrac{\max_d \lambda_d}{\min_d
  \lambda_d}$; $e\in [1,2,5,10]$).  (NB: code available from
\url{https://github.com/mdqyy/simple-gmm-initializations}.)
In total, 30 combinations (out of the 48 possible) are selected, and
further aggregated into three so-called {\em groups}: {\em spherical} (12
models), {\em elliptical} (9 models), {\em elliptical-difficult} (9 models).  Each
GMM is used to generate $D=30$ datasets, yielding a total of 900
datasets, each involving $n=10,000$ points.

We also consider {\em noisy} datasets, namely noisy
spherical/elliptical/elliptical-difficult models.  To generate a noisy
dataset, we generate 9000 points from the noise free model, and add
1000 points drawn uniformly at random in the expanded bounding box of
the 9000 samples.

Summarizing, we consider in the sequel six groups involving 60 models
(30 noise free, 30 noisy), for a total of $D=1800$ datasets.

\paragraphmini{Grid dataset.}
In addition, we include an artificial pathological case with the
\textit{grid dataset}.  In this group, we include data sampled from a
single handcrafted 3-dimensional GMM composed of 27 Gaussians forming
the shape of a 3d cubic grid (Fig.\ref{fig:grid-views}). The rationale
is to highlight situations where the \kgmmseedingGGD initialization
method might be particularly well suited, as the data is generated
using a GMM with easily identifiable but highly intersecting and
eccentric components. With this GMM, we also generate 30 datasets,
each composed of 250 * 27 = 6750 points.

\paragraphmini{Statistics.}
We aim at comparing $N_c=9$ initialization contenders using the
log-likelihood -- Eq. \eqref{eq:log-likelihood}, for every group of
models out of six.  For a given dataset, final log-likelihood values are 
obtained by averaging the results of $R=30$ runs of each initialization method
in order to assess the variance inherent to their randomness.
This yields a vector of final log-likelihood values of size $N_c$ for
each dataset (900 vectors for noise free datasets, 900 for noisy
datasets).
Consider the resulting $D\times N_c$ matrix, with $D=1800$ and
$N_c=9$.  To be able to accumulate results over different datasets
from the same group, we perform min-max scaling over each matrix row,
such that its entries are in $[0,1]$ (with 0 (resp. 1) corresponding
to the worst (resp. best) performing method at each row). These
rescaled values are termed the {\em min-max normalized
  log-likelihoods}.
The matrix columns can then be split into $6$ blocks, each corresponding
to a specific group.  To compare the six groups, we average the
{\em min-max normalized log-likelihood} values of each method on all
datasets of a given block -- resulting in $6 \times N_c$ values in
total.

\subsection{Results}
%%ii-%-%-%-%-%-%-%-%-%-%-%-%-%-%-%-%-%-%-%-%-%-%-%-%-%-%-%-%-%-%-%-%-%-%-%-%-%-%-%

\paragraphmini{Seeding and the final log-likelihood.}
The following observations stand out
(Fig. \ref{fig:graphical-abstract},
Fig.\ref{fig:results_EM_mean_plot}, 
Fig.\ref{fig:grid-results},
SI Table~\ref{tab:results_table_EM} (exact values)).

%% Overall, \kgmmseedingEGDEGC remains the best method,
%% as seen from the improvement of the log likelihood (LL)
%% (Fig. \ref{fig:seeding-time-EM}), with a negligible impact on performance (Fig.\ref{fig:total-time-EM}). 

\sbulem{The zig-zag strategy is state-of-the-art.}  EM combined by
  the classical \kmeanspp seeding, a.k.a \kgmmseedingEGD, is
  outperformed by the variant using twice as many candidates
  (\kgmmseedingEGDtwo), which is itself outperformed by
  \kgmmseedingEGDEGC.  Consistent with \kmeans, this corroborates the
  general efficacy of the zig-zag strategy in increasing the final
  log-likelihood on all classes of datasets.

\sbulem{Log-likelihood (LL) based methods are sensitive to noise.}  The
  comparison between noise free and noisy datasets yields a clear
  separation between methods using the log-likelihood for seed
  selection.  As illustration comparison is that between
  \kgmmseedingEGDEGC and \kgmmseedingEGDEGL-- which differ only
by the metric used to rank candidates.
One the one hand, LL based methods are amongst the best on our noise
free datasets.  On the other hand, these methods appears highly
sensitive to outliers (Fig. \ref{fig:noisy-elliptical-dot5} and
Fig. \ref{fig:noisy-elliptical-sep2}).

\sbulem{Seeding using the Gaussian based distance
is highly effective for mixtures with intersections.}
The case of \kgmmseedingGGD, which uses the $\Dsquare$ strategy
  on the Gaussian distance performs on par with \kgmmseedingEGD for
  generated datasets.  On the grid datasets, it is the best performing
  method overall, and significantly outperforms \kgmmseedingEGD as a
  single pass strategy (Fig.~\ref{fig:grid-results}).  As opposed to
  the likelihood regulated methods, \kgmmseedingGGD incorporates the
  advantages of estimating Gaussian components without suffering from
  the inclusion of noise.

\begin{comment}
The best performing method for this experiment remains
\kgmmseedingEGDEGC. As for \kmeans, we compare the difference in results
between \kgmmseedingEGDEGC and \kgmmseedingEGD, to the difference between
\kgmmseedingEGD and \kgmmseedingEON. The ratios in percentage between
these differences on all 1800 datasets average at 28.94\%, with a
median of 28.26\%.
\end{comment}

\paragraphmini{Running time.}  Similarly to the \kmeans case, the
zig-zag seeding methods are slower that their one pass counterpart
(Fig.~\ref{fig:seeding-time-EM}), but the SSE regulated methods remain
competitive (Fig.\ref{fig:total-time-EM}). Most importantly, the
likelihood regulated passes are 5x to 6x slower than the SSE regulated
passes. Finally, the \kgmmseedingGGD method is around two to three
orders of magnitude slower than all shown methods due to the cost of
estimating Gaussians on each data points (data not shown).

\vspace{\cheatheight}
\begin{figure}[htbp]
\centerline{ \includegraphics[width=.8\linewidth]{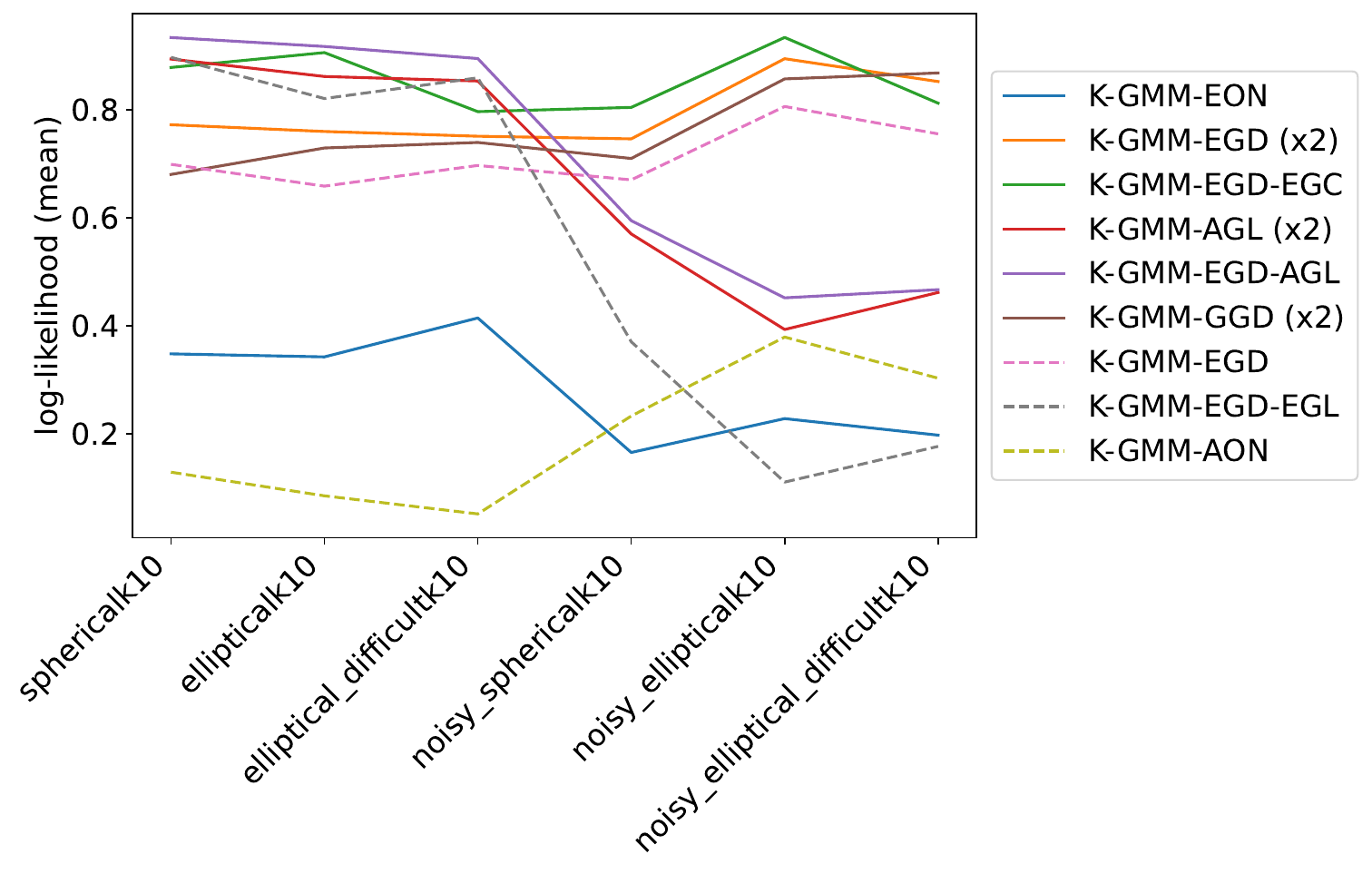}}
\caption{{\bf \kgmm: mean of the  min-max normalized log-likelihood over datasets of each scenario.} 
The larger the log-likelihood, the better. See text for details.}
\label{fig:results_EM_mean_plot}
\end{figure}

\ifLONG
\begin{figure}[htbp]
\centering
\includegraphics[width=.7\linewidth]{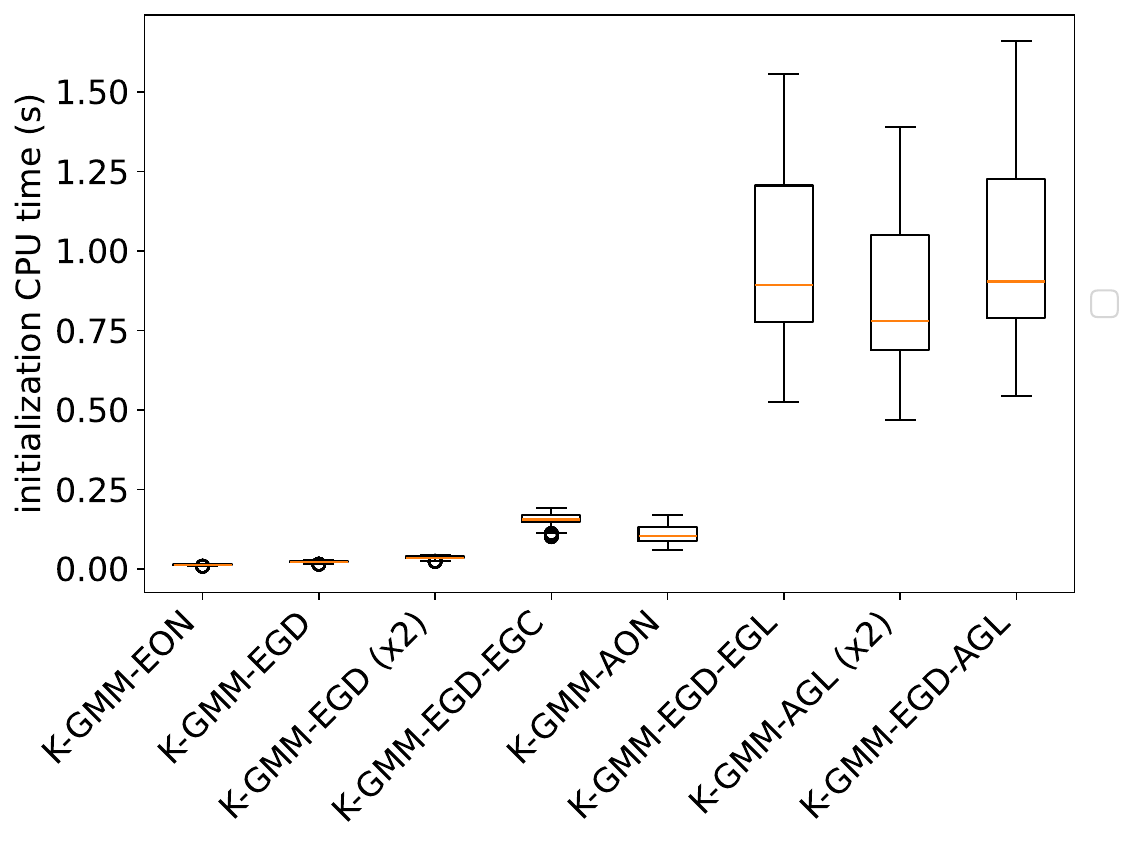}
\caption{{\bf \kgmm: CPU time in seconds for each seeding method.}}
\label{fig:seeding-time-EM}
\end{figure}
\fi

\begin{comment}
 Let $\delta$
the LL increase between \kgmmseedingEGDEGC and \kgmmseedingEGD, and
$\Delta$ the same between
\kgmmseedingEGD
and \kgmmseedingEON. The min, median, mean (stdev), and max values of $\delta
/\Delta$ are: -61.64, 0.28, 0.28 (3.47), 54.06 (Fig.\ref{fig:graphical-abstract}). The standard deviation of 3.47 indicates that the mix and max values are outliers, in fact 91.78 \% of values fall in the range $[-0.5, 3]$. 
\end{comment}

\section{Outlook}
\label{sec:outlook}
%%i%%%%%%%%%%%%%%%%%%%%%%%%%%%%%%%%%%%%%%%%%%%%%%%%%%%%%%%%%%%%%%%%%%%%%%%%%%%%%%%

Clustering is a fundamental problem, and \kmeans is a fundamental
approach to it.  In search for efficient and provably correct
solutions, the smart seeding approach of \kmeanspp play a central
role.  Re-seeding methods based on local searches and multi-swaps
recently underwent important developments, both in theory and in
practice.  We improve on these in several ways.

On the design side, while recent re-seeding methods have used
distances to the data points, we use  distances to the
centroids of the clusters induced by the centers, in the spirit of the
early works of local searches. Also, our methods are particularly
simple and do not require any elaborate data structure to maintain
nearest neighbors of the seeds under scrutiny during re-selection.
Our methods achieve SOTA performance but currently
lack theoretical guarantees, specifically in terms of a constant
approximation factor (CFA). This situation parallels that of
\kmeansppg, which has been the preferred practical method since 2007
despite the absence of theoretical analysis until 2023, and since 2023
despite having a CFA worse than that of \kmeanspp.
Our experiments also shed light on subtle properties of \kmeans often
overlooked, including the (lack of) correlations between the SSE upon
seeding and the final SSE, the variance reduction phenomena observed
in iterative seeding methods and the sensitivity of the final SSE to the
pool size for greedy methods.
\medskip

Practically, we anticipate that our best seeding methods will become one of
the standard seeding technique(s).
On the theoretical side, the analysis of our  methods raises challenging questions. 
The first one is the role of the metric and its coherence with the functional eventually optimized
by Lloyd iterations for \kmeans, or EM iterations for \kgmm.
The second relates to the ordering along which seeds are being opted
out. In a manner akin to simulated annealing, which lowers the
temperature strategy, the better performances of our {\em zig-zag}
strategy advocate a re-seeding processing seeds with {\em lower}
variance first. Yet, this intuition has to be formally established.

\toblue
\noindent{\bf Acknowledgments.} 
This work has been supported by the French government, through the 3IA C\^ote d’Azur Investments 
(ANR-19-P3IA-0002), and the ANR project Innuendo (ANR-23-CE45-0019).
\toblack

%%\bibliographystyle{alpha}
%% included
\clearpage
\bibliographystyle{unsrt}
%\bibliography{\wmybib/biogeom,\wmybib/mcs,\wmybib/abs,\wmybib/abs-sub}

%% For journals like Proteins etc
%% \renewcommand{\figurename}{Figure S\hspace{-.05cm}~}
%% \renewcommand{\tablename}{Table S~}
%% \setcounter{figure}{0}
%% \setcounter{table}{0}
%% \makeatletter %% package algorithm
%% \renewcommand{\ALG@name}{Algorithm S\hspace{-.1cm}~}
%% \makeatother
\beginSI
\newpage

%% included
%% \input{seeding-SI-v2.tex}
\section{Supporting information: theory}
%%i%%%%%%%%%%%%%%%%%%%%%%%%%%%%%%%%%%%%%%%%%%%%%%%%%%%%%%%%%%%%%%%%%%%%%%%%%%%%%%%

\subsection{\kmeans}
%%i%%%%%%%%%%%%%%%%%%%%%%%%%%%%%%%%%%%%%%%%%%%%%%%%%%%%%%%%%%%%%%%%%%%%%%%%%%%%%%%
\begin{algorithm}
\begin{algorithmic}[1]
\Procedure{\seedingEONEON}{$data, K$}
\Statex
\State{$centers \gets$ \seedingkmeanspp(data, K)}
\LComment{Reselect centers in reverse order}
\For{$k \gets K$ to $1$}
\State{Delete $centers[k]$}
\State{Choose $new\_c_k$ with the $\Dsquare$ strategy}
\State{Insert $new\_c_k$ in $centers$ at position $k$}    
\EndFor
\EndProcedure
\end{algorithmic}
\caption{{\bf \seedingEONEON.}}
\label{alg:SeedingEONEON}
\end{algorithm}

\subsection{EM for Gaussian Mixtures}
\label{sec:EM-equations}
%%ii-%-%-%-%-%-%-%-%-%-%-%-%-%-%-%-%-%-%-%-%-%-%-%-%-%-%-%-%-%-%-%-%-%-%-%-%-%-%-%

\subsubsection{The  \meanstogmm  and \meanstosphgmm algorithms}

Consider a dataset and a hard partition of this dataset into clusters.

The algorithm \ref{alg:means2gmm}~\citep{blomer2016adaptive} converts
this partition into a Gaussian mixture model. An interesting
observation is that it is often beneficial to estimate isotropic
initial components instead of anisotropic ones.

\begin{algorithm}
\begin{algorithmic}[1]
\Procedure{\meanstogmm}{$X, {\mu_1, ..., \mu_K}$}

\Statex
\State{Derive partition ${C_1, ..., C_K}$ of X by assigning each point $x_i \in X$ to its closest mean}
\LComment{Build GMM components}
\For{$k \gets 1$ to $K$}
\State{$\mu_k = 1 / |C_k| \ \sum_{x\in C_k}x$}
\State{$w_k = |C_k|\  /\  |X|$}
\State{$\Sigma_k = 1 / |C_k| \ \sum_{x\in C_k}(x-\mu_k)\transpose{(x-\mu_k)}$}\label{line:aniso}
\State{If $\Sigma_k$ is not positive definite, take $\Sigma_k = 1/(d\size{C_k}) \sum_{x\in C_k} \vvnorm{x-\mu_k}{2}\idmatrix{d})$}\label{line:iso}
\State{If $\Sigma_k$ is still not positive definite, take $\Sigma_k = \idmatrix{d}$}
\EndFor
\EndProcedure
\end{algorithmic}
\caption{{\bf The classical \meanstogmm algorithm.}
The variant \meanstosphgmm consists of changing the
full anisotropic estimation of  line \ref{line:aniso} by the isotropic estimation of line \ref{line:iso}.
\citep{blomer2016adaptive}.
}
\label{alg:means2gmm}
\end{algorithm}

%% Given a data set $X\subset \Rd$, the {\em Maximum Likelihood
%%   Estimation} (MLE) is to find a K-GMM model $\Theta$ maximizing the likelihood
%% $\calL(\Theta) = \prod_{x\in X} \normalD{x}{\Theta}$.
%% %%
%% Given an initial model, EM can be used to refine it -- see
%% Section \ref{sec:EM-equations}. 

\subsubsection{The  E and M steps}
In the following, we recall the  EM algorithm to fit a GMM.
The algorithm involves two steps.

\paragraph{The E-step.} Given the functions at iteration $t$, one computes the  responsibility 
of the gaussian $g^{(t)}_j$ for the sample point $x_i$
\begin{equation}
\emcres[t]{ij} = \frac{ \emcw[t]{j} g^{(t)}_j(x_i|\Theta^{(t)}_j)}{\sum_{k=1}^M \emcw[t]{k} g^{(t)}_k(x_i|\Theta^{(t)}_k)}.
\end{equation}
The sum of responsibilities associated to one component then read as
\begin{align}
\emcsum[t]{j} = \sum_{i=1}^N \emcres[t]{ij}
\end{align}
%% Bishop p 432
It may be noted \citep{bishop2006pattern}(Chapter 9) that the weight
$\emcw[t]{j}$ is the prior probability for the sample $x_i$ to be
generated by the j-th component; the responsibility $\emcres[t]{ij}$is
the corresponding posterior probability.

\paragraph{The M-step.}  Re-estimate the parameters of $g^{(t+1)}_j$  using the maximum likelihood:
\begin{gather*}
\label{eq:gtp1}
\emcw[t+1]{j} = \frac{\emcsum[t]{j}}{N},\\
\emcmu[t+1]{j}  = \frac{1}{\emcsum[t]{j}} \sum_{i=1}^n  \emcres[t]{ij} x_i,\\
\Sigma_j^{(t+1)}  = \frac{1}{\emcsum[t]{j}} \sum_{i=1}^n \emcres[t]{ij}  (x_i-\emcmu[t+1]{j}) \transpose{(x_i-\emcmu[t+1]{j})}.
\end{gather*}

\paragraph{Convergence.}
Consider the log likelihood for the $n$ samples, that is
\begin{equation}
\label{eq:log-likelihood}
\loglik{X}{\Theta} = 
\ln \normalD{X}{\Theta} = 
\sum_{i=1,\dots,n} 
\ln \bigl( \sum_k w_k \normalD{x_i}{\mu_k, \Sigma_k} \bigr).
\end{equation}
One checks the convergence of the mixture parameters, or of the likelihood.

\paragraph{Numerics.}
In the E-step of the EM algorithm, the evaluation of the GMM needed
to compute responsibilities induces a risk of underflow. This is due
to the evaluation of singular components on datapoints for which they
have no responsibility, resulting in pdf values that tend to 0. \\ To
solve this problem, we adapt the E-step by computing the
responsibilities using only the logarithmic scale. We first compute
the log-pdfs of singular components for individual unnormalized
responsibilities. To obtain normalized responsibilities, we must
compute the log pdf of the whole mixture model. 

This computation requires a summation of the pdf values of components, which cannot be explicitly obtained without losing the logarithmic scale. We avoid this problem by performing the logsumexp trick (https://en.wikipedia.org/wiki/LogSumExp) using the log-pdfs of individual components, allowing us to obtain the log-pdf of the whole mixture model while staying in logarithmic scale throughout.

The logsumexp operation consists in the following:

\begin{equation}
logsumexp(x_1, ..., x_n) = \log(\sum_{i=1}^n \exp(x_i))  
\end{equation}

Applied on the log-pdfs of individual components, this allows us to
obtain the log-pdf of the mixture model, but loses the logarithmic
scale. Therefore, we apply the logsumexp trick by using the following
equivalent to the logsumexp operation:
\begin{equation}
logsumexp(x_1, ..., x_n) = x^* + \log(\sum_{i=1}^n \exp(x_i - x^*))  
\end{equation}

This equivalent allows us to shift the values in the exponent by an
arbitrary constant. We can then set $x^* = \max\{x_1, ..., x_n\}$, to
ensure the largest exponentiated term is equal to $\exp(0)=1$,
avoiding the risk of underflow on the result of the logsumexp
operation.

\subsection{$D^2_G$ with Gaussian distance}
\label{sec:kgmmseedingGGD}

%%ii-%-%-%-%-%-%-%-%-%-%-%-%-%-%-%-%-%-%-%-%-%-%-%-%-%-%-%-%-%-%-%-%-%-%-%-%-%-%-%

\paragraph{Multivariate Gaussians and associated distance}
The parameter set of a multi-dimensional Gaussian is denoted $\Theta=(\mu, \Sigma)$; it has dimension
$d + d(d+1)/2 = d(d+3)/2$.
The set of positive semidefinite matrices, to which covariance
matrices belong, is denoted $\PSD$.

The density of a multivariate Gaussian reads as:
\begin{equation}
g(x|\Theta) = 
\frac{1}{\sqrt{(2\pi)^d \determinant{\Sigma}}}  \expL{-\frac{1}{2} \transpose{(x-\mu)} \Sigma^{-1} (x-\mu)}.
\end{equation}

Let $\Sigma_1$ and $\Sigma_2$ be two PSD matrices; let $\muonetwo = \mu_1-\mu_2$,
and let $\Sigma_{12} = (\Sigma_1+\Sigma_2)/2$.
Consider the generalized eigenvalue problem (GEP) $\Sigma_1 V = \lambda \Sigma_2 V$,
with $V$ the column matrix of the generalized eigenvectors.
The Riemannian metric for PSD reads as~\citep{forstner2003metric}:
\begin{equation}
\distPSD{\Sigma_1, \Sigma_2} = \big( \sum_{i=1}^d \log^2 \lambda_j\bigr)^{1/2}.
\end{equation}
Using this, one defines the following distance between Gaussian distributions~\citep{abou2010designing}:
\begin{equation}
\label{eq:dist-gauss}
\distGauss{g_1, g_2} = 
\bigl( \transpose{\muonetwo} \Sigma^{-1} \muonetwo \bigr)^{1/2}   
+ \bigl( \sum_{k=1}^d \ln^2 \lambda_k \bigr)^{1/2}.
\end{equation}

\begin{comment}

%% \citep{bonnabel2009riemannian}
%%       title={Riemannian Metric and Geometric Mean for Positive Semidefinite Matrices of Fixed Rank},
\begin{remark}
The set of PSD matrices $\PSD$ has dimension $d(d+1)/2$, has can be
seen from the following counting argument on eigenvectors and
eigenvalues of the matrix:
\begin{itemize}
\item To define $d-1$ principal directions as unit vectors (the last one being implicitly defined),
one need $(d-1)\times(d-1) - \binom{d-1}{2} = d(d-1)/2$ parameters. The first terms
counts the individual coordinates of the unit vectors, while the second one imposes 
the pairwise orthogonality constraints between these unit vectors.
\item To define the eigenvalues, one needs $d$ values.
\item Whence in total $d(d-1)/2 +d =  d(d+1)/2$ parameters.
\end{itemize}
\end{remark}

\begin{remark}
The distance of Eq.  \ref{eq:dist-gauss} generalizes the {\em
  parameter distance} used in ~\citep{moitra2010settling} for 1D
Gaussians, which is simply the L1 distance between the parameters
(mean and variance).
%%
Under suitable assumptions, there is a correspondence 
between $\distGauss$ and  the {\em statistical} distance  $\distGS$ defined by
\begin{equation}
\distGS{g1, g2} \int_{\Rd} \fabs{f(x)-g(x)} dx.
\end{equation}
See also the references and the discussion in \citep{abou2010designing}.
\end{remark}
\end{comment}

\paragraph{\kgmmseedingGGD: details.}
Using the previous gaussian distance we propose the \kgmmseedingGGD
seeding algorithm. This is a modification of the \kgmmseedingEGD
algorithm, where the \kmeanspp seeding method is used on locally
estimated gaussians at each data point, instead of the data points
themselves. In other words, this method aims at sampling seeds from
the datapoints by considering the shape of the gaussian components
that would result from selecting them.  This local estimation of
gaussians is done as follows:

\begin{itemize}
\item At each data point $x_j$, compute average distance $\hat d_j$ to $L$-nearest neighbors $\{x^{(j)}_l\}_{l=1}^L$ with the following equation. Considering K components will be estimated in the following EM algorithm, we set $L$ to be equal to $N / K$ .

\begin{equation}
\label{eq:di}
\hat d_j = \frac{1}{L}\sum_{l=1}^L ||x_l^{(j)} - x_j||.
\end{equation}

\item For each pair of points $x_i$ and $x_j$, compute the local distance weighted responsibility of point $x_j$ for point $x_i$ with the following equation. These responsibilities are designed to correspond to the evaluation on the point $x_i$ of an isotropic gaussian with variance $\hat d_j$, centered on the data point $x_j$.

\begin{gather*}
LG_j(x_i) = \frac{1}{(\hat d_j)^D (2\pi)^{D/2}} \exp(- \frac{1}{2} \frac{||x_i - x_j||^2}{(\hat d_j)^2})\\
\tilde{r}_{ij} = \frac{LG_j(x_i)}{\sum_{k=1}^N LG_k(x_i)}.
\end{gather*}

\item At each data point, compute local gaussians $G_j$ with the M-step update rules of the EM algorithm (\ref{eq:gtp1}), using the local responsibilities $\tilde r_{ij}$.
\end{itemize}

This process provides a set $G$ of locally estimated gaussians of size
$N$, one gaussian for each datapoint. It is then followed by a
selection of $K$ gaussians among $G$ with the \kmeanspp algorithm
using the gaussian distance of Eq.\ref{eq:dist-gauss}. Finally, we
select the $K$ data points from which the $K$ selected gaussians were
obtained as starting centers for the EM iterations, completing the
seed selection.

\section{Supporting information: results}
%%i%%%%%%%%%%%%%%%%%%%%%%%%%%%%%%%%%%%%%%%%%%%%%%%%%%%%%%%%%%%%%%%%%%%%%%%%%%%%%%%

\subsection{\kmeans}
%%ii-%-%-%-%-%-%-%-%-%-%-%-%-%-%-%-%-%-%-%-%-%-%-%-%-%-%-%-%-%-%-%-%-%-%-%-%-%-%-%

\begin{figure}[htb]% or !htb or H
\centerline{ 
\includegraphics[width=.5\linewidth]{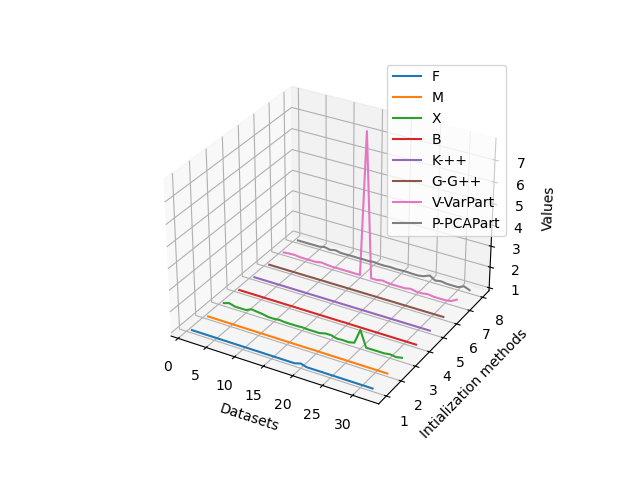}\hfill
\includegraphics[width=.5\linewidth]{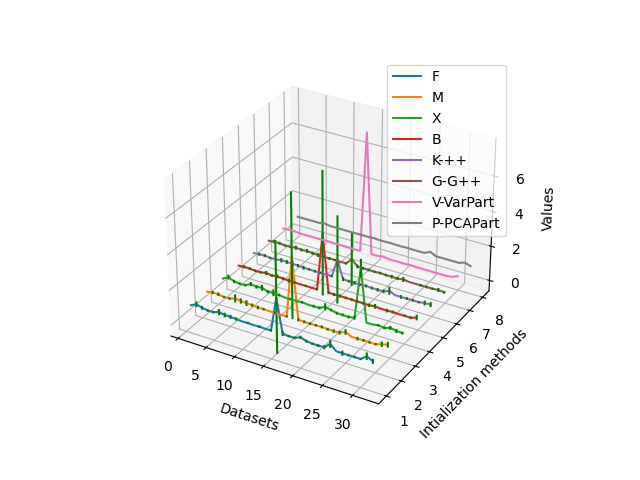}}
\caption{{\bf Plot of Table 2 from \citep{celebi2013comparative}, on 32 datasets.}
{\bf (Left)} Minimum values scaled by the minimum for a dataset
{\bf (Right)} Mean values scaled by the minimum for a dataset. The error bars
correspond to mean values $\pm$ the std deviation.}
 \label{fig:celebi-table2} 
\end{figure} 

\begin{figure}[htb]% or !htb or H
\centerline{ \includegraphics[width=.5\linewidth]{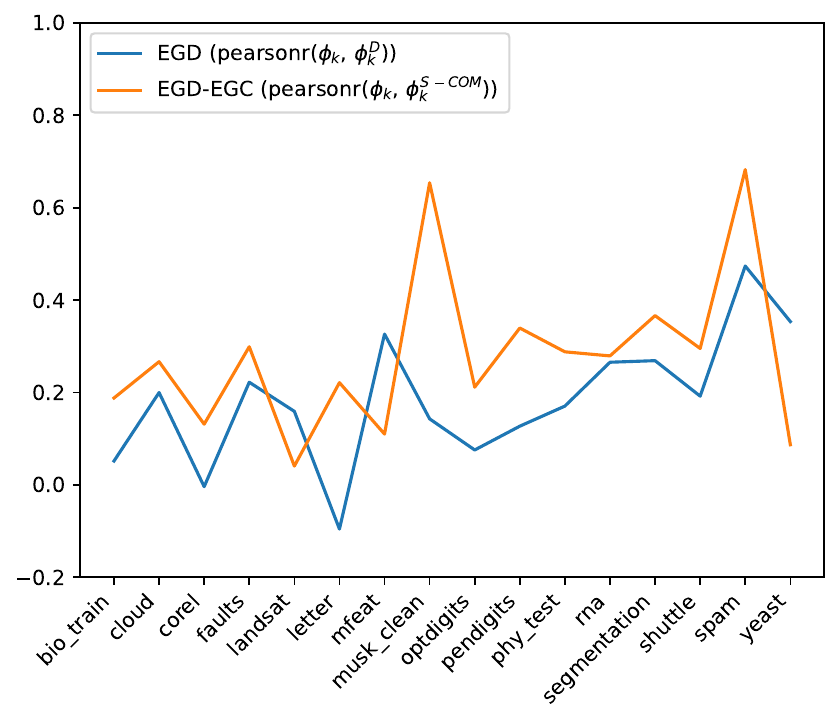}\hfill
 \includegraphics[width=.5\linewidth]{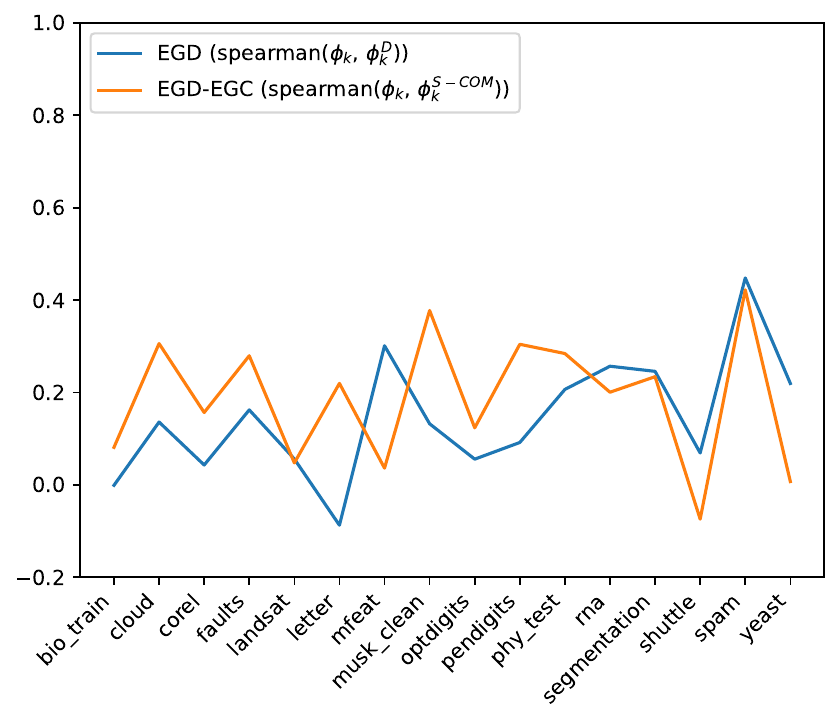}}
\caption{{\bf \kmeans: Correlations 
between 
$(\kmeansfun, \kmeansfunS)$
and ($\kmeansfun, \kmeansfunCOM)$
	on the datasets from  \citep{celebi2013comparative}}. Correlations are computed from the values of 150 repeats of \kmeans, using as initialization \kmeansppg for $\kmeansfunS$ and \seedingEGDEGC for $\kmeansfunCOM$. }
\label{fig:correlations} 
\end{figure}

\begin{figure}[htb]
\begin{center}
\begin{tabular}{cc}
\includegraphics[width=.45\textwidth]{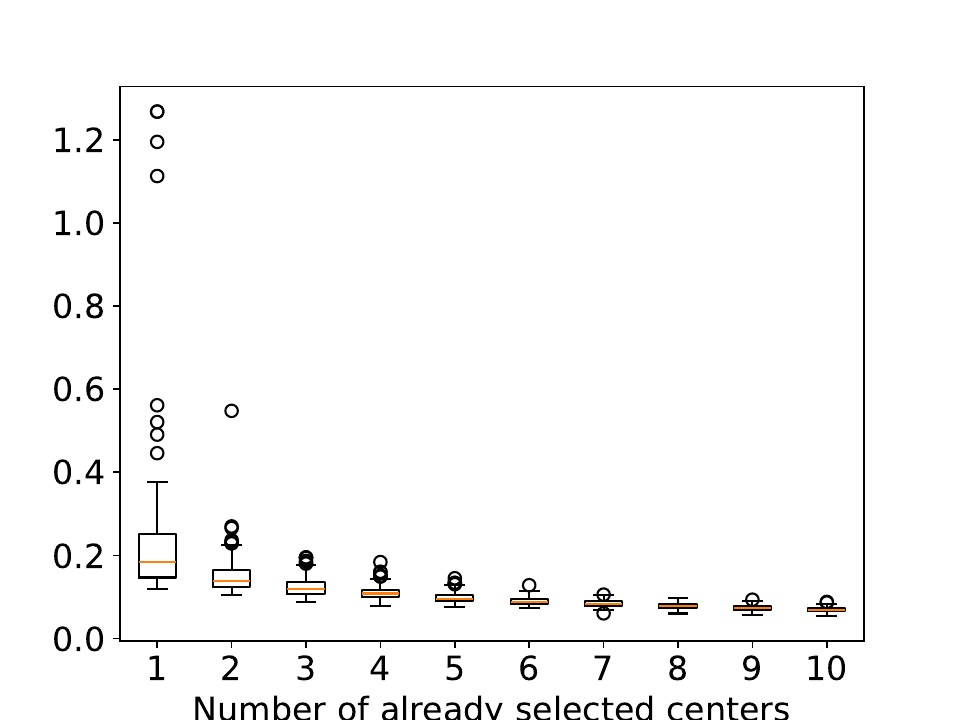} &
\includegraphics[width=.45\textwidth]{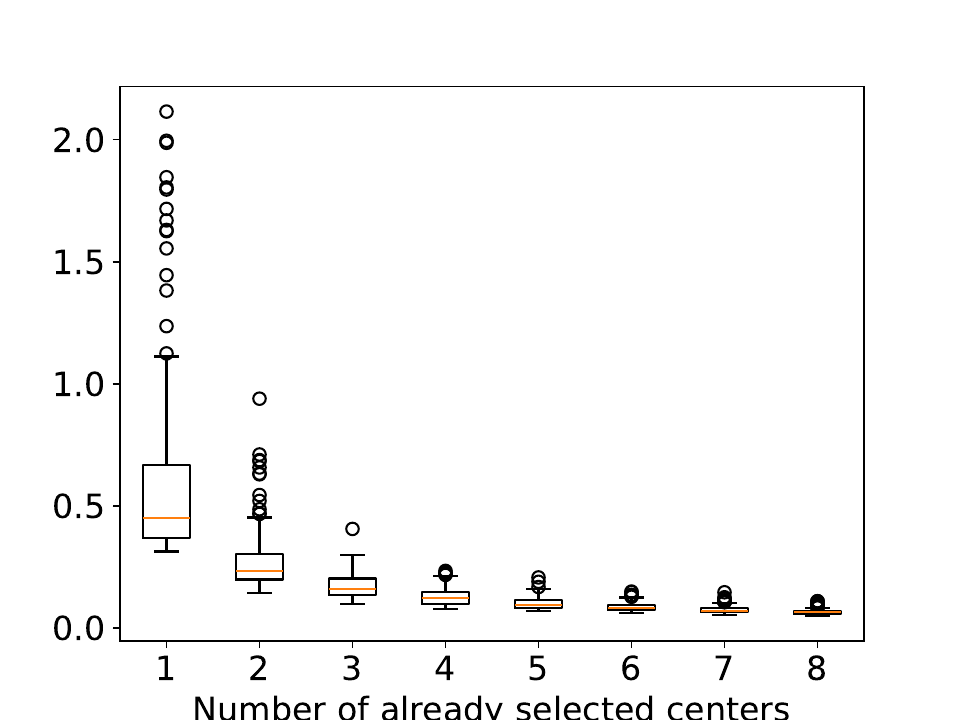}\\
yeast & cloud\\
\includegraphics[width=.45\textwidth]{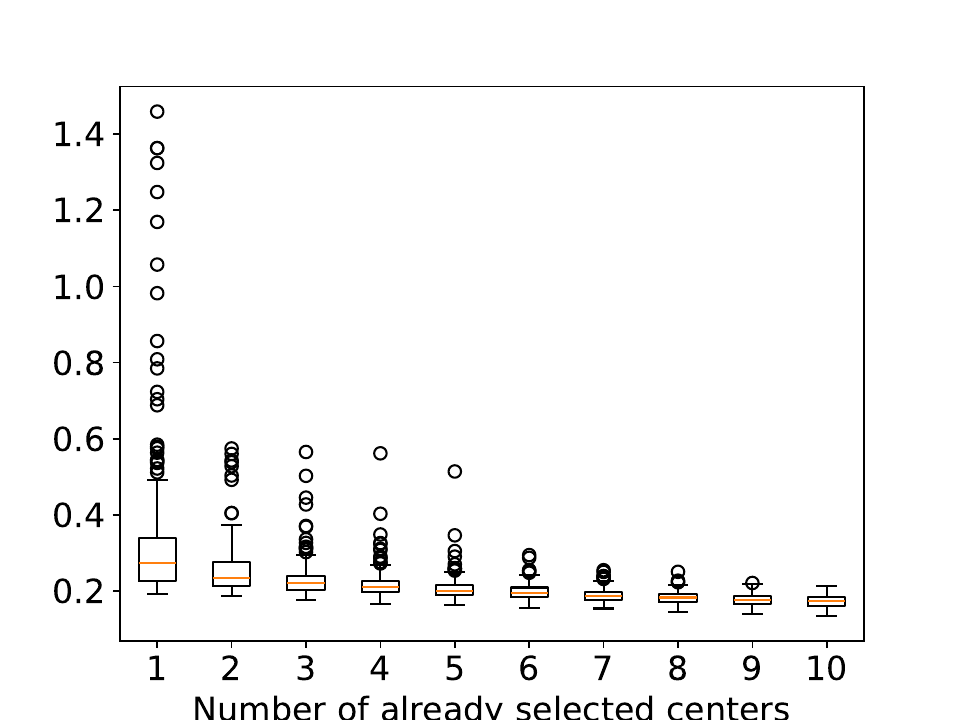}&
\includegraphics[width=.45\textwidth]{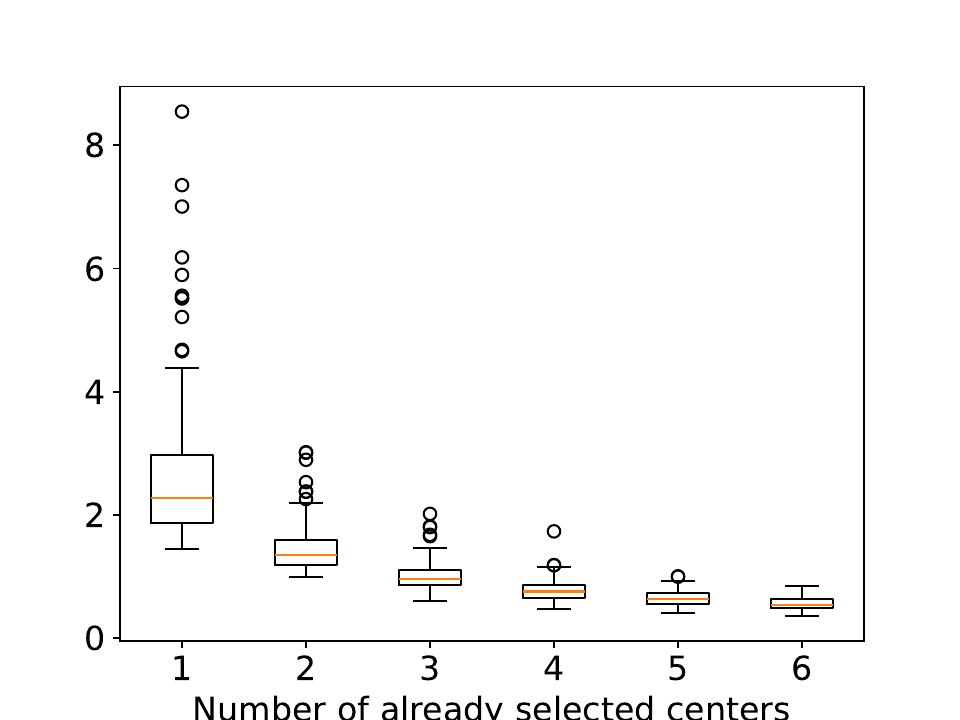}\\
spam & landsat\\
\end{tabular}
\end{center}
	\caption{{\bf \kmeans: boxplot of the mean square distance $\Dsquaremean$ (each sample to its nearest seed) along the seeding selection process -- $k \in 1..K$.} Statistics over 150 repeats of \kmeanspp on 
each dataset.}
\label{fig:d2-boxplot}
\end{figure}

\begin{figure}[htb]
\includegraphics[width=\textwidth]{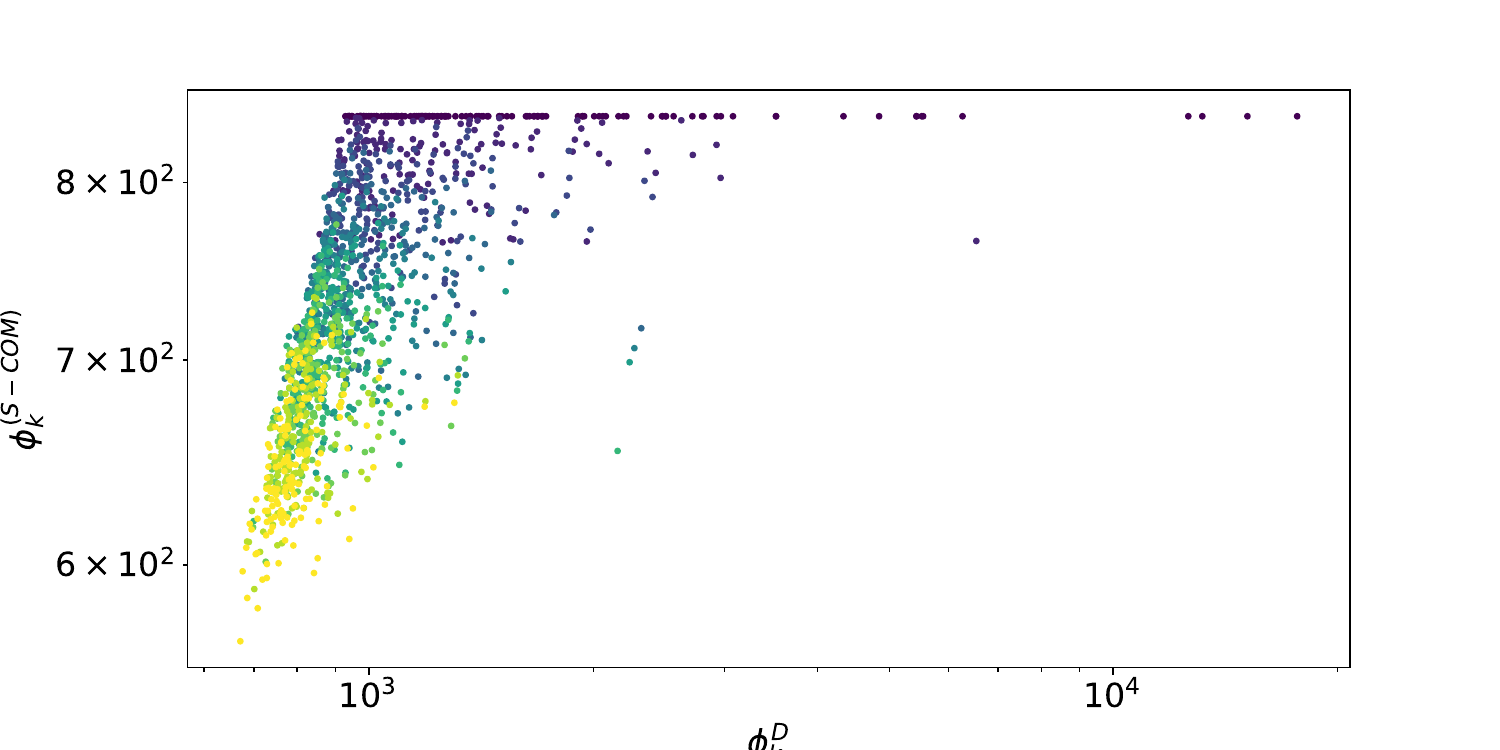}
\caption{{\bf \kmeans: scatterplot of the $\kmeansfunS$ and $\kmeansfunCOM$ values along the seeding selection process for each $k \in 1..K$.}  Statistics over 150 repeats on \textit{spam} dataset. The darker the dot, the earlier in the selection process the values were measured.}
\label{fig:greedy-metrics-scatter}
\end{figure}

\begin{figure}[htbp]
\centerline{ \includegraphics[width=\linewidth]{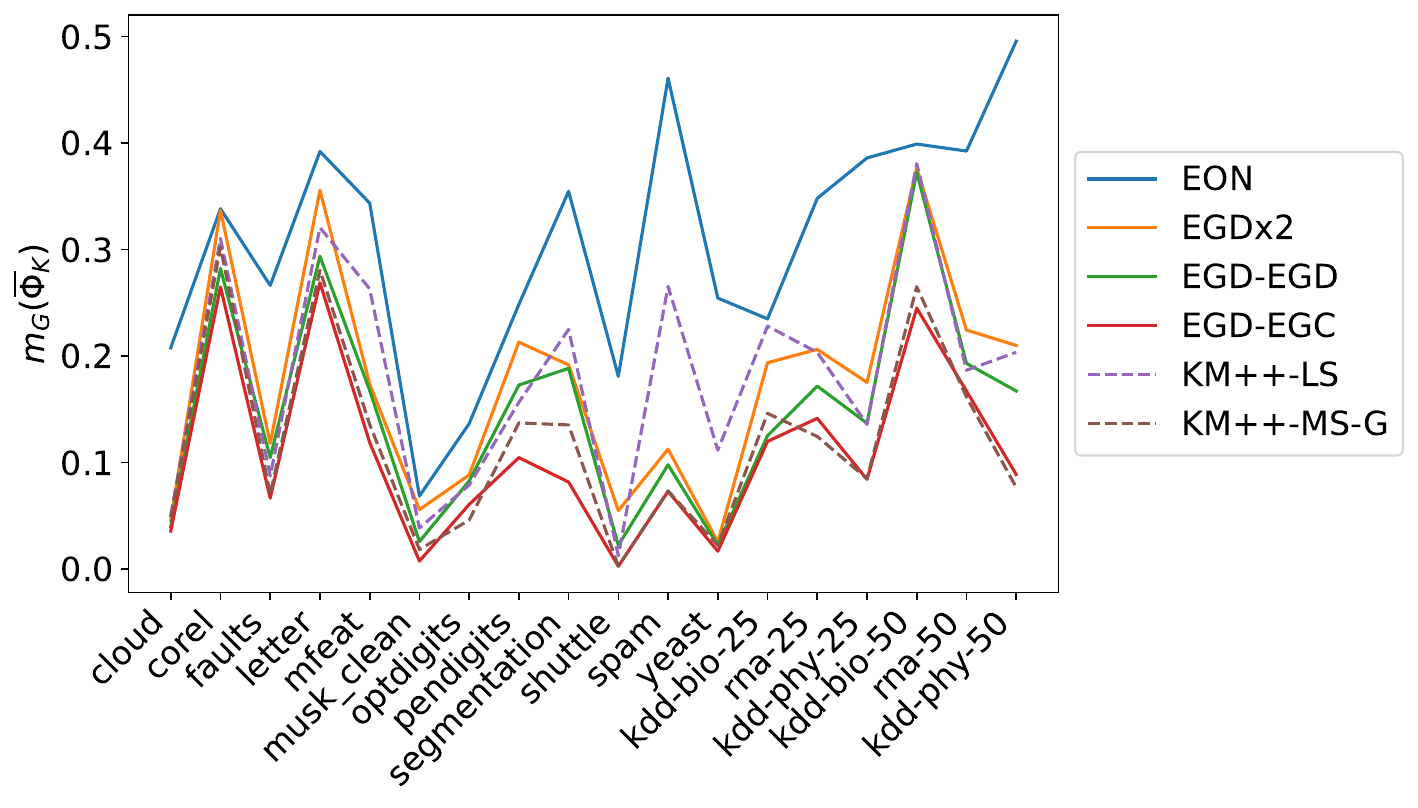}}
\caption{{\bf \kmeans: min-max normalized $\mmG$ -- Eq. \eqref{eq:min-max-glob}  as a function
of the seeding method.}}
%%
%This scaling allows for comparing the method's
%erformance relative to the dataset.}
\label{fig:kmeansfunc-means2}
\end{figure}

\ifLONG
\else
\begin{figure}[htbp]
\centerline{\includegraphics[width=\linewidth]{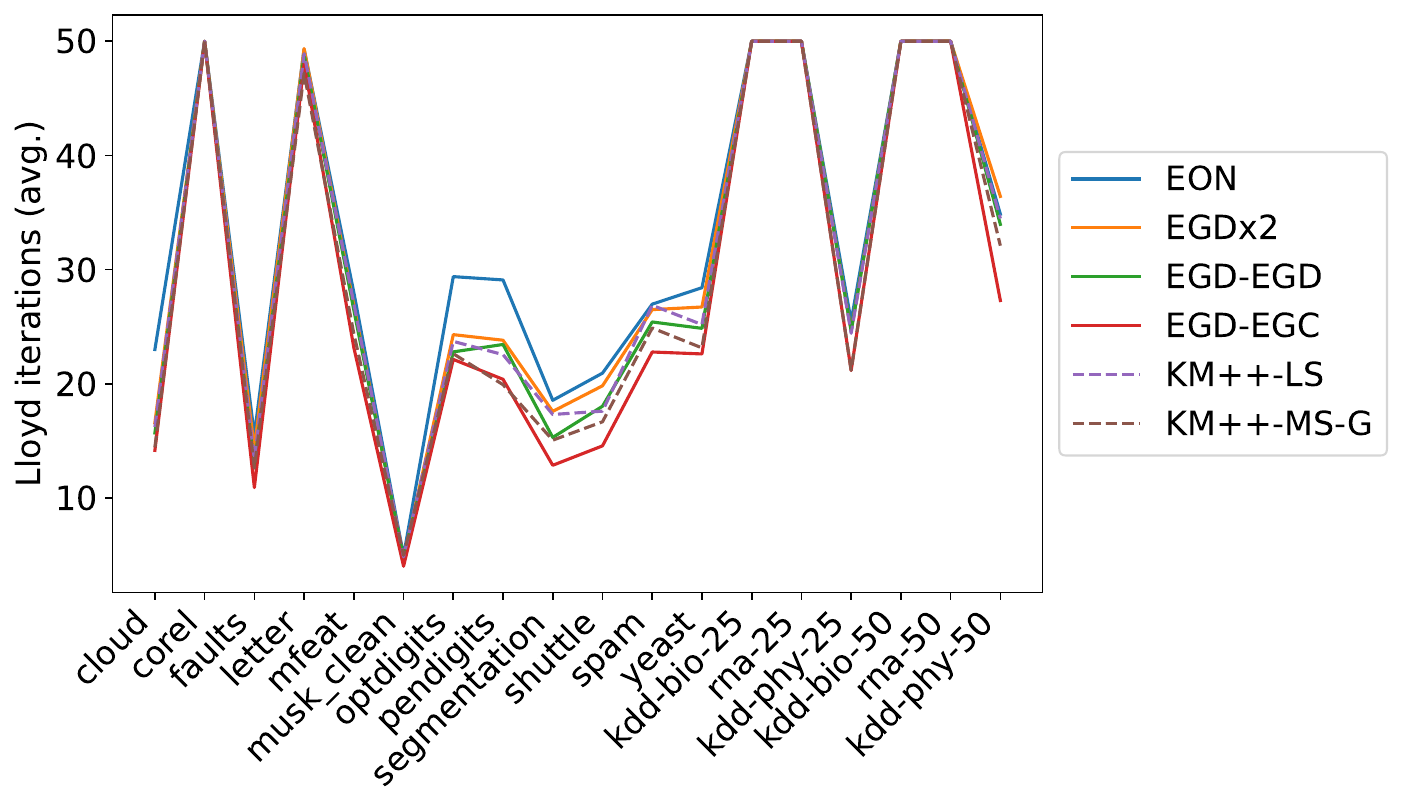}}
\caption{{\bf \kmeans: incidence of the seeding method on the average number of Lloyd iterations.}}
\label{fig:num-Lloyd-iterations-kmeans}
\end{figure}
\fi

\begin{figure}[htbp]
\centerline{\includegraphics[width=\linewidth]{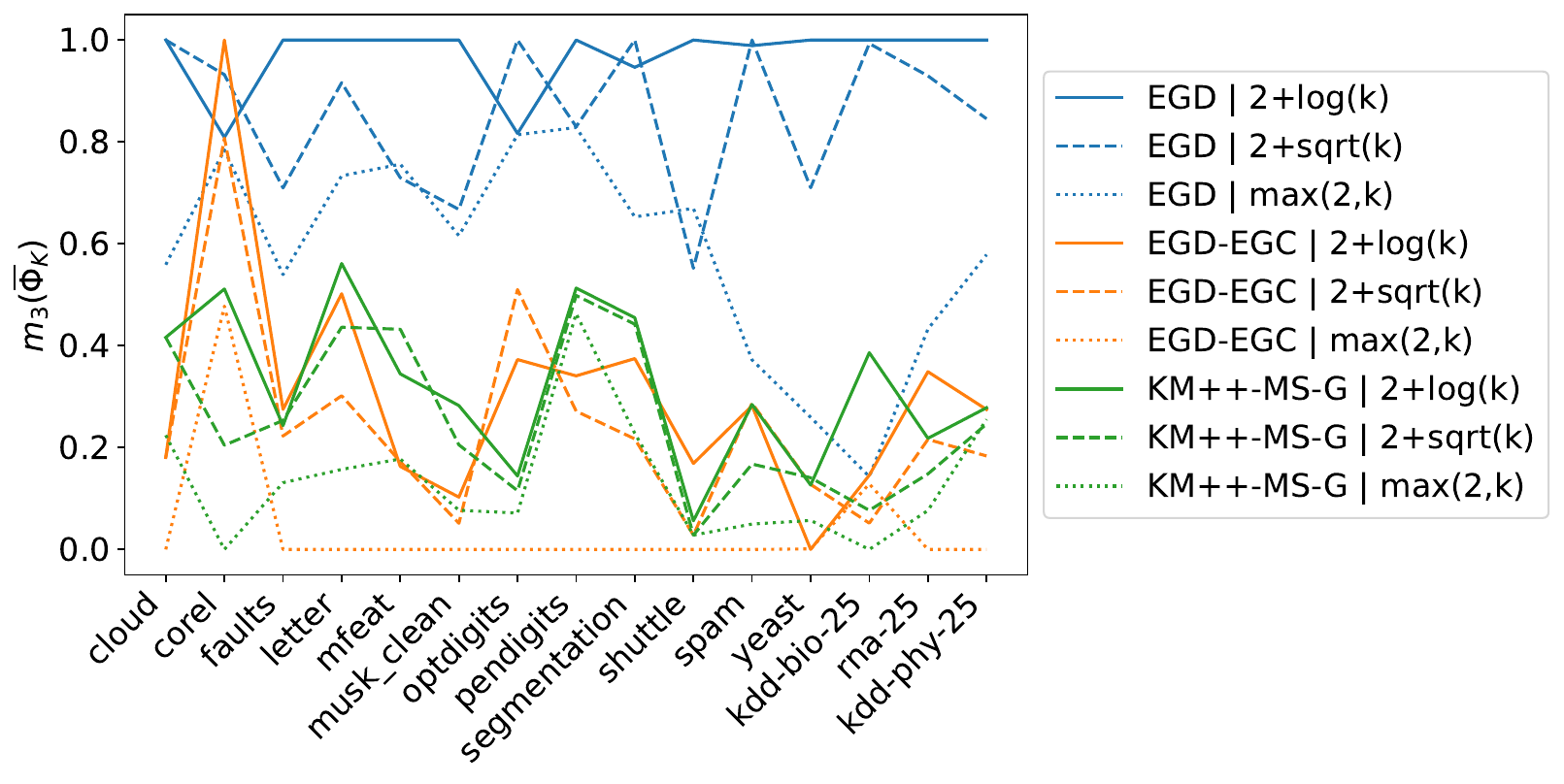}}
\caption{{\bf \kmeans: min-max normalized $\mmM$ -- Eq. \eqref{eq:min-max-mean}, 
as a function of the seeding method and the candidate pool size.}}
\label{fig:parameter_kmeans_func}
\end{figure}

\begin{figure}[htbp]
\centerline{\includegraphics[width=\linewidth]{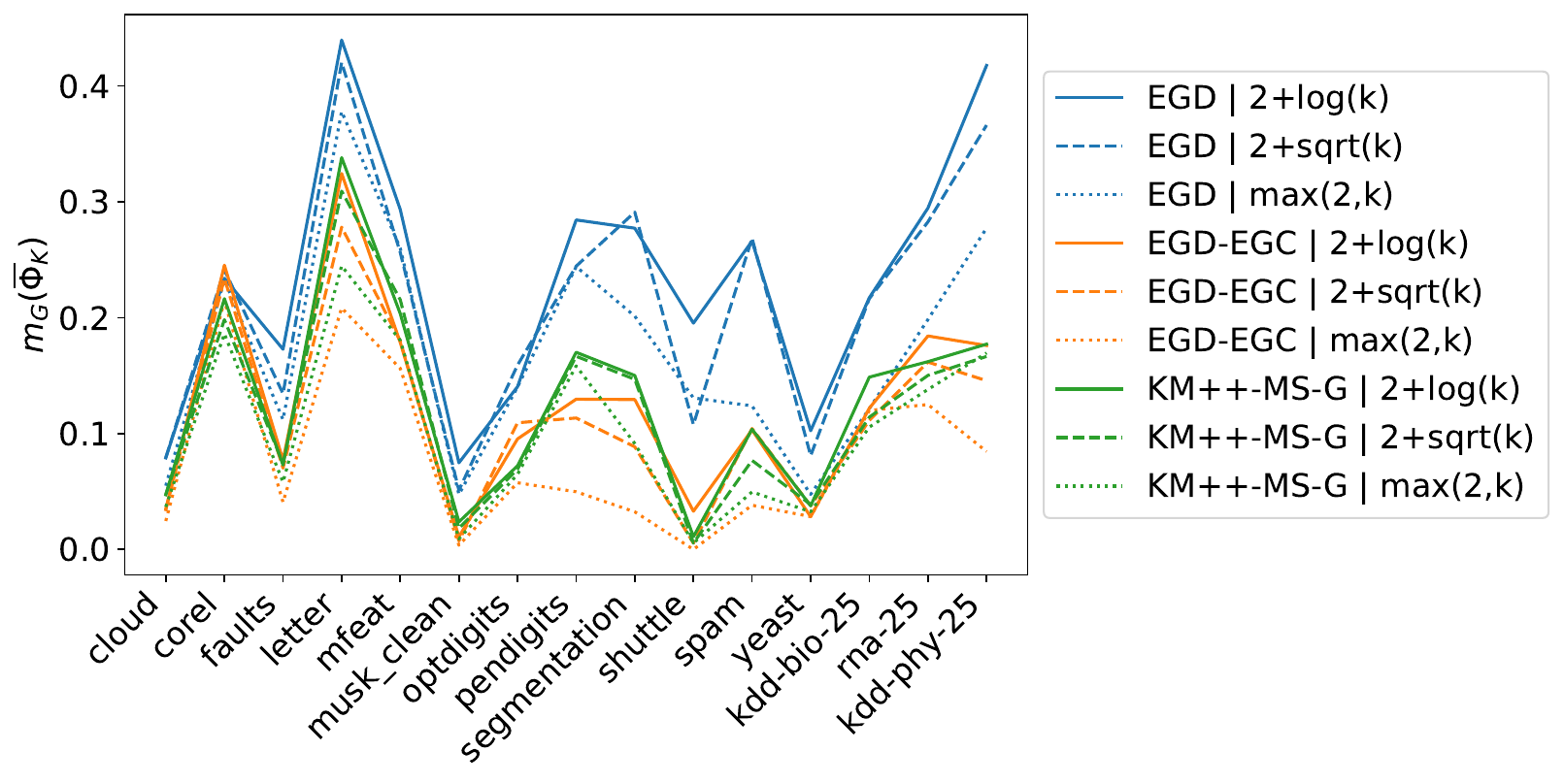}}
\caption{{\bf \kmeans: min-max normalized $\mmG$ -- Eq. \eqref{eq:min-max-glob}, 
as a function of the seeding method and the candidate pool size.}} 
\label{fig:parameter_kmeans_func2}
\end{figure}

\begin{figure}[htbp]
\centerline{\includegraphics[width=\linewidth]{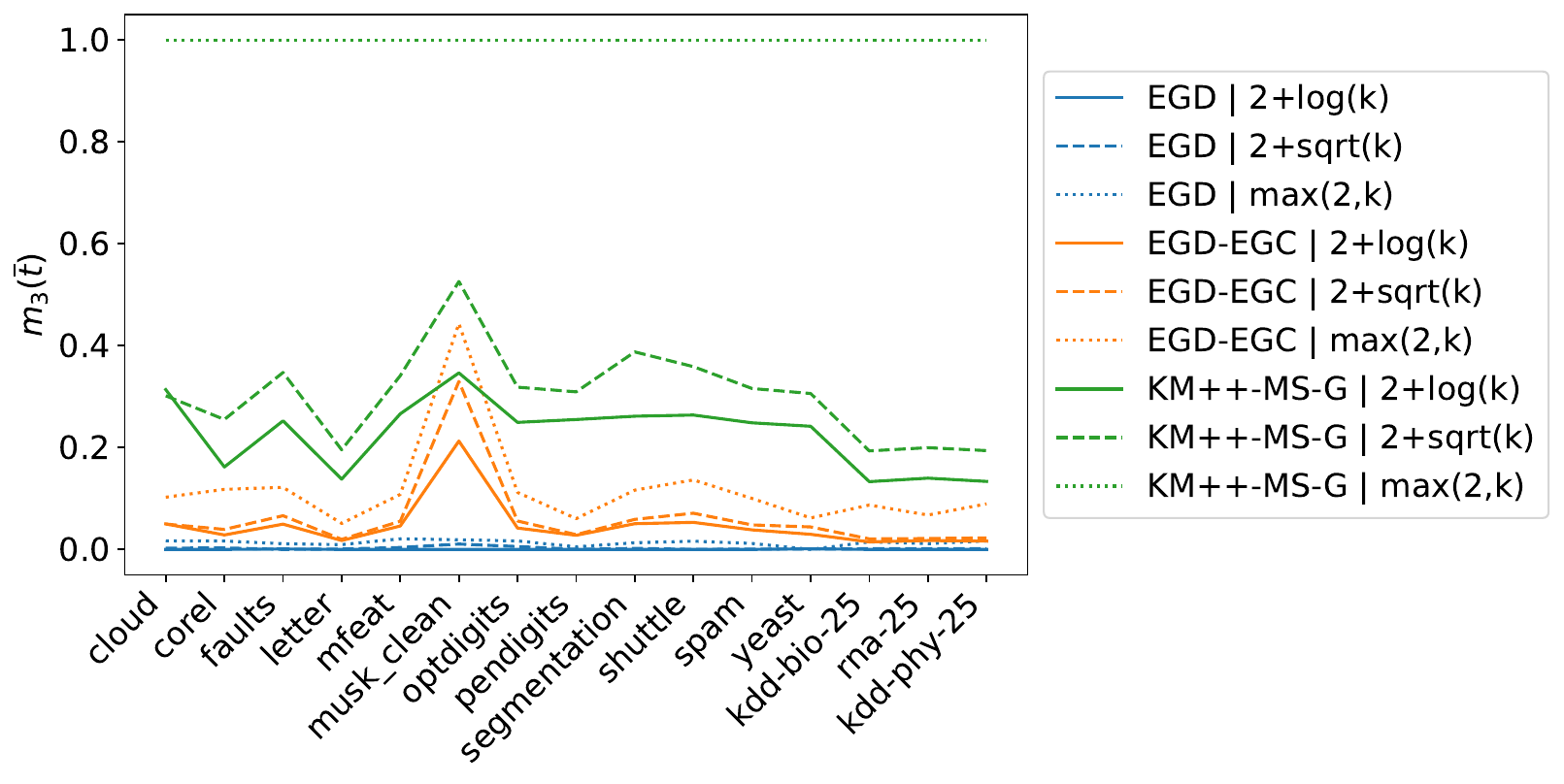}}
\caption{{\bf \kmeans: min-max normalized CPU time $\mmM{\bar{t}}$
as a function of the seeding method and the candidate pool size.}}
\label{fig:parameter_seeding_time_kmeans}
\end{figure}

\begin{figure}[htbp]
\centerline{\includegraphics[width=\linewidth]{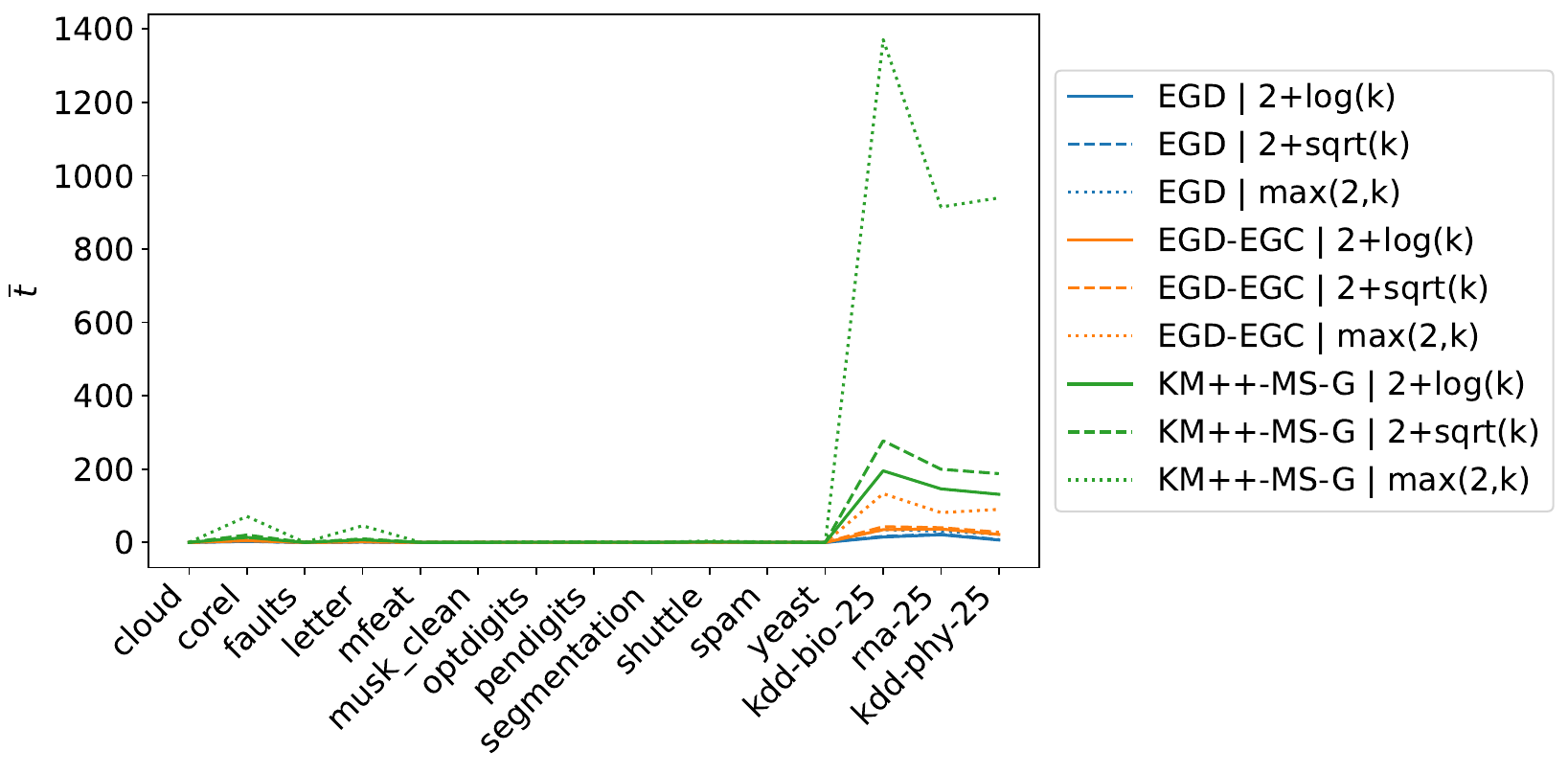}}
\caption{{\bf \kmeans: Raw CPU time (in seconds) as a function of the seeding method and the candidate pool size.}}
\label{fig:parameter_seeding_time_kmeans2}
\end{figure}

\begin{landscape}
\begin{table}
\center
\tiny
\begin{tabular}{|c|c|c|c|c|c|c|c|}
\hline
 & \tiny & \tiny EON & \tiny EGDx2 & \tiny EGD-EGD & \tiny EGD-EGC & \tiny KM++-LS & \tiny KM++-MS-G \\\hline
cloud & \tiny min & \tiny 123.52 & \tiny 123.52 & \tiny 123.52 & \tiny 123.52 & \tiny 123.52 & \tiny 123.52\\
 & \tiny mean & \tiny 130.90 ± 10.29 & \tiny 125.13 ± 2.06 & \tiny 124.93 ± 1.75 & \tiny \bf{124.78} ± 1.36 & \tiny 126.18 ± 6.22 & \tiny 125.25 ± 2.13\\\hline
corel & \tiny min & \tiny 10060.20 & \tiny 10058.70 & \tiny 10057.90 & \tiny 10058.60 & \tiny 10058.20 & \tiny 10057.30\\
 & \tiny mean & \tiny 10095.31 ± 21.07 & \tiny 10095.15 ± 22.22 & \tiny 10089.01 ± 22.33 & \tiny \bf{10087.06} ± 18.82 & \tiny 10095.81 ± 25.61 & \tiny 10091.28 ± 22.49\\\hline
faults & \tiny min & \tiny 1260.81 & \tiny 1260.74 & \tiny 1260.75 & \tiny 1260.74 & \tiny 1260.76 & \tiny 1260.75\\
 & \tiny mean & \tiny 1318.25 ± 48.02 & \tiny 1286.21 ± 25.95 & \tiny 1283.32 ± 23.15 & \tiny \bf{1275.09} ± 15.21 & \tiny 1287.80 ± 30.52 & \tiny 1275.91 ± 16.48\\\hline
letter & \tiny min & \tiny 2723.84 & \tiny 2721.15 & \tiny 2715.48 & \tiny 2718.34 & \tiny 2719.92 & \tiny 2715.45\\
 & \tiny mean & \tiny 2756.12 ± 19.12 & \tiny 2752.34 ± 17.83 & \tiny 2745.91 ± 15.16 & \tiny \bf{2743.34} ± 12.88 & \tiny 2754.10 ± 17.61 & \tiny 2744.54 ± 14.39\\\hline
mfeat & \tiny min & \tiny 3128.00 & \tiny 3126.49 & \tiny 3127.66 & \tiny 3128.07 & \tiny 3128.64 & \tiny 3127.65\\
 & \tiny mean & \tiny 3171.67 ± 32.60 & \tiny 3149.32 ± 18.61 & \tiny 3148.54 ± 18.64 & \tiny \bf{3142.10} ± 12.47 & \tiny 3165.01 ± 28.88 & \tiny 3144.34 ± 16.01\\\hline
musk-clean & \tiny min & \tiny 36372.70 & \tiny 36372.70 & \tiny 36372.70 & \tiny 36372.70 & \tiny 36372.70 & \tiny 36372.70\\
 & \tiny mean & \tiny 37086.86 ± 1319.74 & \tiny 36953.47 ± 1288.51 & \tiny 36639.47 ± 661.18 & \tiny \bf{36448.92} ± 373.40 & \tiny 36677.58 ± 698.57 & \tiny 36563.25 ± 571.65\\\hline
optdigits & \tiny min & \tiny 14559.20 & \tiny 14559.20 & \tiny 14559.20 & \tiny 14559.20 & \tiny 14559.20 & \tiny 14559.20\\
 & \tiny mean & \tiny 14708.80 ± 207.68 & \tiny 14655.94 ± 116.61 & \tiny 14649.75 ± 121.71 & \tiny 14625.54 ± 80.60 & \tiny 14685.01 ± 158.61 & \tiny \bf{14609.40} ± 85.33\\\hline
pendigits & \tiny min & \tiny 4930.15 & \tiny 4930.15 & \tiny 4930.15 & \tiny 4930.15 & \tiny 4930.15 & \tiny 4930.15\\
 & \tiny mean & \tiny 5073.43 ± 109.36 & \tiny 5052.98 ± 81.65 & \tiny 5029.72 ± 67.03 & \tiny \bf{4990.43} ± 68.30 & \tiny 5047.08 ± 92.85 & \tiny 5009.22 ± 52.44\\\hline
segmentation & \tiny min & \tiny 386.98 & \tiny 386.98 & \tiny 386.98 & \tiny 386.98 & \tiny 386.98 & \tiny 386.98\\
 & \tiny mean & \tiny 410.17 ± 16.20 & \tiny 399.51 ± 12.64 & \tiny 399.29 ± 12.07 & \tiny \bf{392.31} ± 8.57 & \tiny 401.62 ± 12.63 & \tiny 395.82 ± 11.54\\\hline
shuttle & \tiny min & \tiny 234.98 & \tiny 234.98 & \tiny 234.98 & \tiny 234.98 & \tiny 234.98 & \tiny 234.98\\
 & \tiny mean & \tiny 263.56 ± 35.45 & \tiny 243.63 ± 17.91 & \tiny 238.48 ± 11.15 & \tiny 235.37 ± 3.88 & \tiny 238.48 ± 11.15 & \tiny \bf{235.37} ± 3.88\\\hline
spam & \tiny min & \tiny 526.08 & \tiny 525.21 & \tiny 524.80 & \tiny 524.78 & \tiny 525.04 & \tiny 524.80\\
 & \tiny mean & \tiny 566.65 ± 18.80 & \tiny 534.99 ± 8.64 & \tiny 533.66 ± 8.72 & \tiny \bf{531.39} ± 7.07 & \tiny 556.24 ± 16.64 & \tiny 531.43 ± 7.37\\\hline
yeast & \tiny min & \tiny 58.33 & \tiny 58.28 & \tiny 58.28 & \tiny 58.28 & \tiny 58.28 & \tiny 58.28\\
 & \tiny mean & \tiny 63.59 ± 5.38 & \tiny 58.80 ± 0.45 & \tiny 58.74 ± 0.47 & \tiny \bf{58.62} ± 0.37 & \tiny 62.31 ± 5.04 & \tiny 58.74 ± 0.37\\\hline
kdd-bio-25 & \tiny min & \tiny 28534.60 & \tiny 28546.60 & \tiny 28540.90 & \tiny 28534.70 & \tiny 28548.80 & \tiny 28544.50\\
 & \tiny mean & \tiny 28643.24 ± 132.45 & \tiny 28624.17 ± 111.63 & \tiny 28592.46 ± 65.25 & \tiny \bf{28589.99} ± 46.53 & \tiny 28666.10 ± 151.53 & \tiny 28602.17 ± 83.51\\\hline
kdd-phy-25 & \tiny min & \tiny 137276.00 & \tiny 137367.00 & \tiny 137301.00 & \tiny 136745.00 & \tiny 138237.00 & \tiny 136783.00\\
 & \tiny mean & \tiny 144905.69 ± 3654.05 & \tiny 140447.92 ± 1438.83 & \tiny 139638.87 ± 1239.44 & \tiny \bf{138522.96} ± 919.62 & \tiny 142497.14 ± 2843.84 & \tiny 138536.81 ± 792.31\\\hline
rna-25 & \tiny min & \tiny 16690.40 & \tiny 16661.20 & \tiny 16654.50 & \tiny 16652.40 & \tiny 16668.50 & \tiny 16653.50\\
 & \tiny mean & \tiny 16941.96 ± 154.55 & \tiny 16823.94 ± 105.86 & \tiny 16795.23 ± 93.95 & \tiny 16770.02 ± 85.54 & \tiny 16877.44 ± 129.90 & \tiny \bf{16755.85} ± 74.52\\\hline
kdd-bio-50 & \tiny min & \tiny 26609.40 & \tiny 26616.10 & \tiny 26614.60 & \tiny 26609.40 & \tiny 26606.40 & \tiny 26613.20\\
 & \tiny mean & \tiny 26659.91 ± 32.01 & \tiny 26657.06 ± 28.52 & \tiny 26656.56 ± 26.37 & \tiny \bf{26640.41} ± 15.77 & \tiny 26662.84 ± 30.81 & \tiny 26642.95 ± 19.02\\\hline
kdd-phy-50 & \tiny min & \tiny 116132.00 & \tiny 113349.00 & \tiny 112627.00 & \tiny 112559.00 & \tiny 115498.00 & \tiny 112487.00\\
 & \tiny mean & \tiny 119015.89 ± 1958.53 & \tiny 115251.90 ± 849.30 & \tiny 114688.41 ± 790.79 & \tiny 113655.36 ± 536.03 & \tiny 117959.49 ± 1662.75 & \tiny \bf{113496.50} ± 483.98\\\hline
rna-50 & \tiny min & \tiny 12568.40 & \tiny 12552.20 & \tiny 12551.40 & \tiny 12517.40 & \tiny 12546.40 & \tiny 12530.20\\
 & \tiny mean & \tiny 12719.97 ± 105.27 & \tiny 12633.19 ± 48.48 & \tiny 12617.03 ± 41.81 & \tiny 12603.64 ± 32.61 & \tiny 12677.79 ± 90.82 & \tiny \bf{12600.74} ± 37.07\\\hline
\end{tabular}
 \caption{\kmeans seeding: $\kmeansfun$ values for each seeding method}
 \label{tab:results_table_kmeans} 
 \end{table}
\end{landscape}

\subsection{\kgmm}
%%ii-%-%-%-%-%-%-%-%-%-%-%-%-%-%-%-%-%-%-%-%-%-%-%-%-%-%-%-%-%-%-%-%-%-%-%-%-%-%-%

\begin{figure}[htb]% or !htb or H
\centerline{ \includegraphics[width=.5\linewidth]{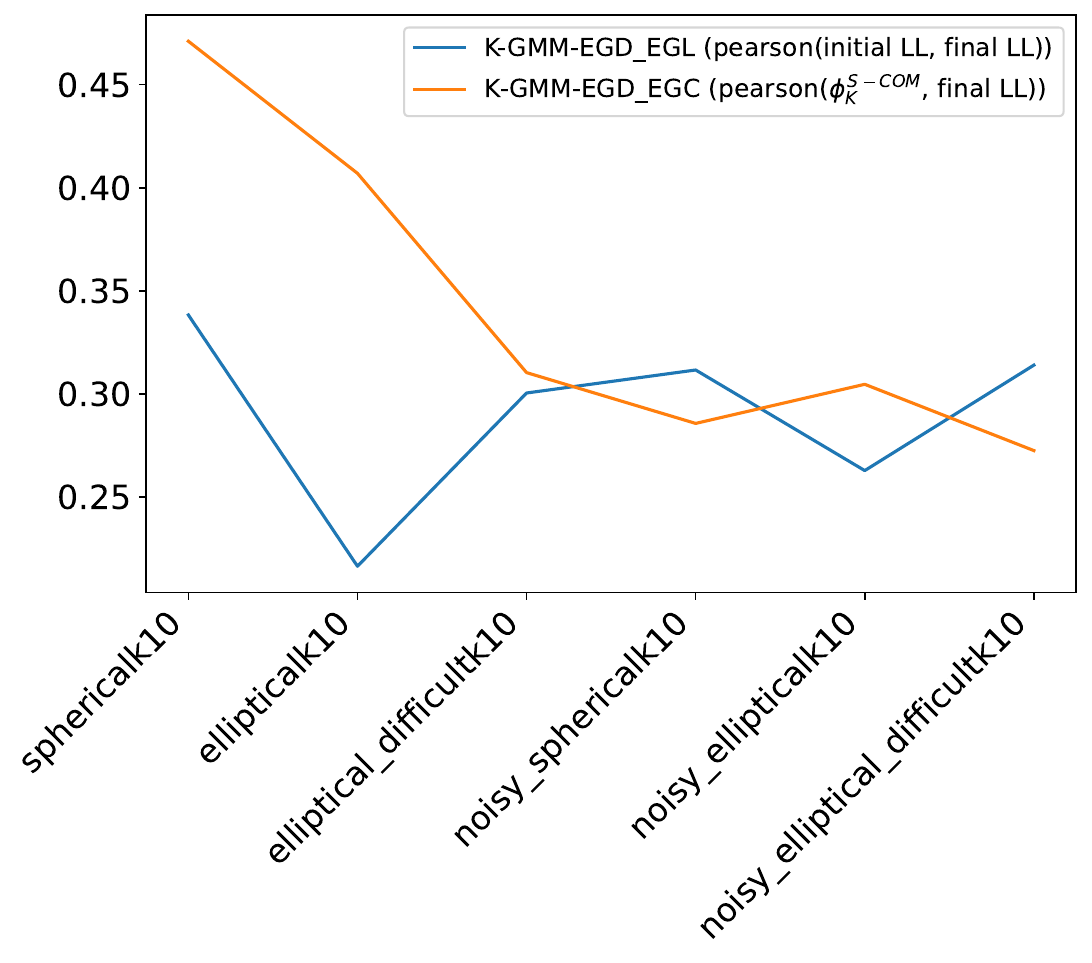}\hfill
\includegraphics[width=.5\linewidth]{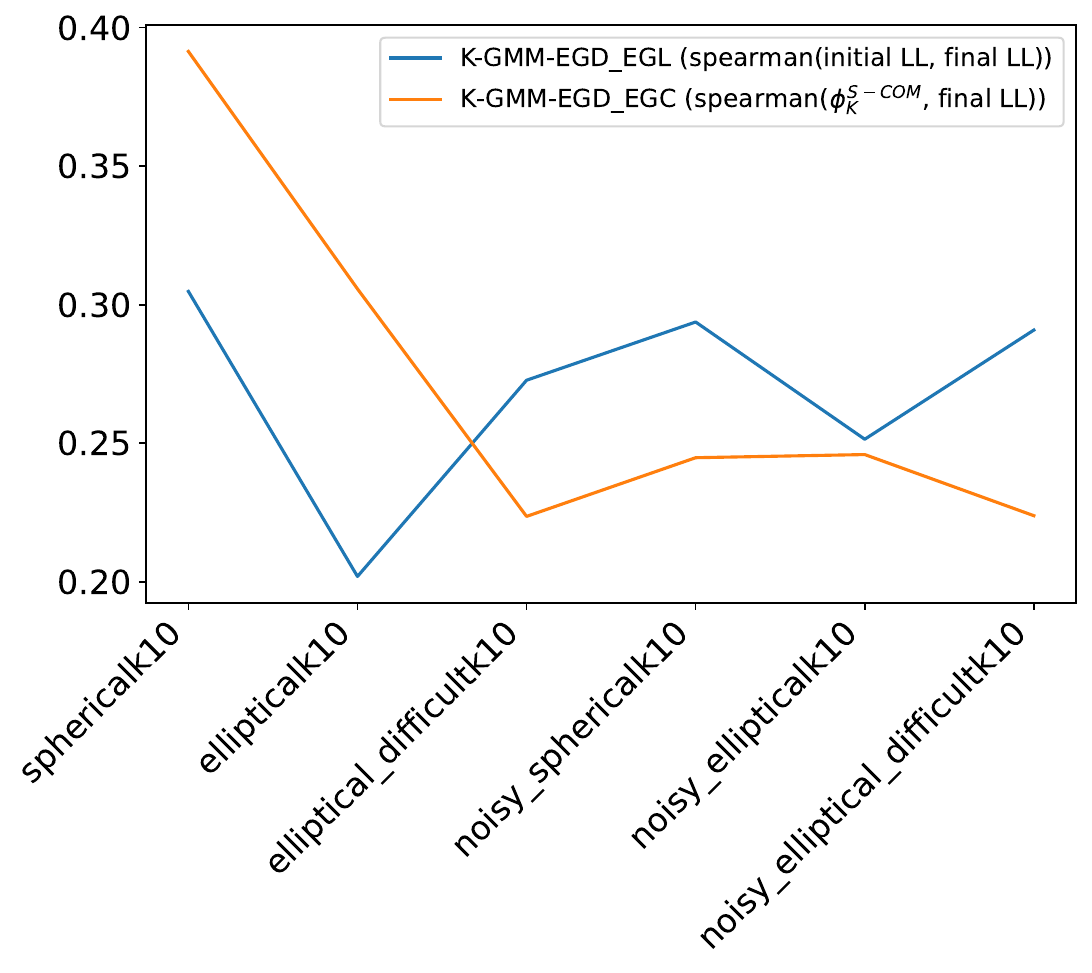}}
\caption{{\bf \kgmm : Correlations (in absolute value)
 between (initial log-likelihood, final log-likelihood)
and ($\kmeansfunCOM$, final log-likelihood)
on the generated datasets from \citep{blomer2013simple}.} Correlations are computed from the values of 30 repeats of EM per dataset, using as initialization \kgmmseedingEGDEGC for log-likelihood and \kgmmseedingEGDEGC for $\kmeansfunCOM$.}
\label{fig:correlations-EM}
\end{figure}

\begin{figure}[htb]
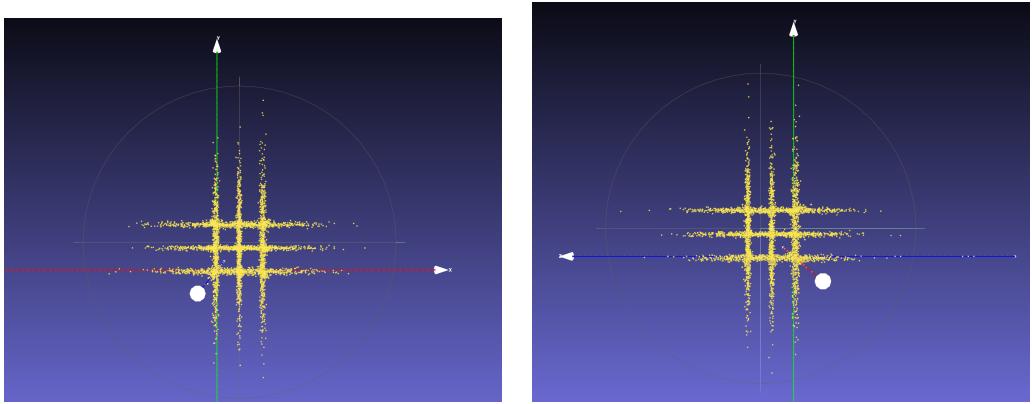

\begin{center}
\begin{tabular}{cc}
\includegraphics[width=.4\linewidth]{./grid1} &   \includegraphics[width=.4\linewidth]{./grid2}
\end{tabular}
\end{center}
\caption{{\bf \kgmm : Top and side views of a 3D point cloud generated by a mixture of 27 anisotropic Gaussian distributions.}}
\label{fig:grid-views}
\end{figure}

\begin{figure}[htb]
\centering
\includegraphics[width=.9\linewidth]{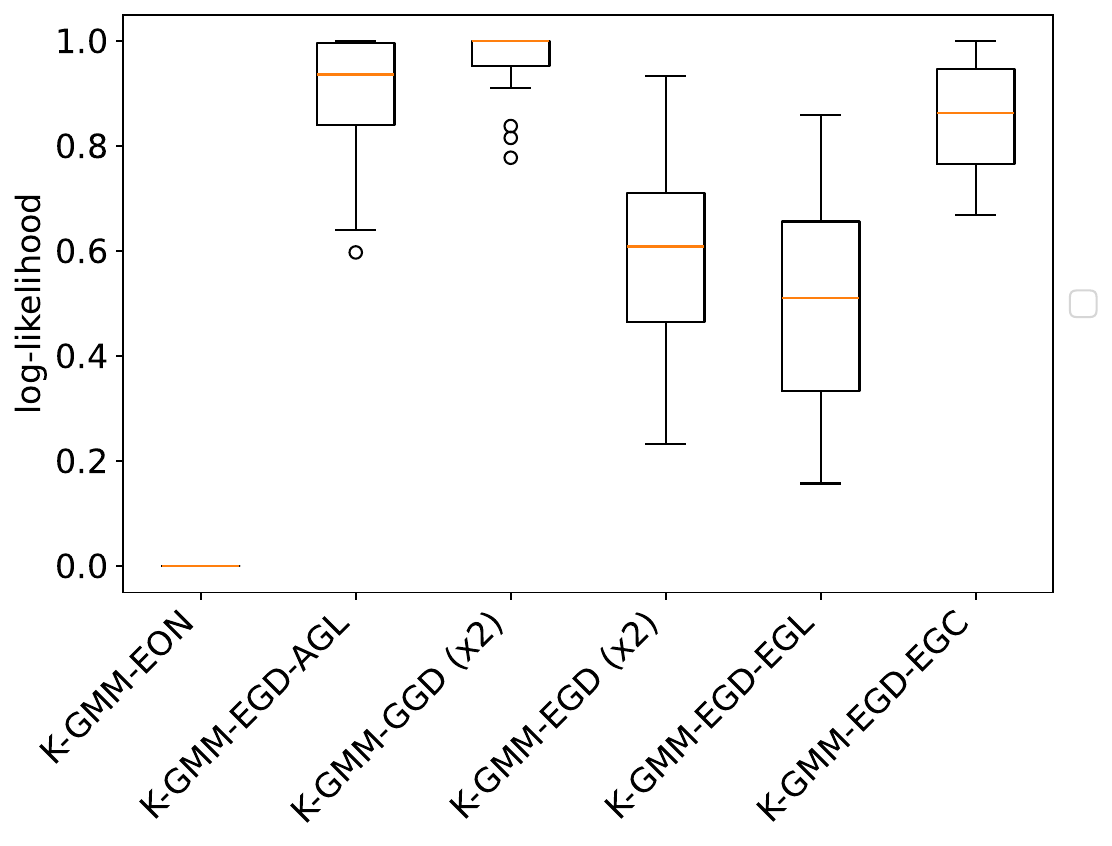}
\caption{{\bf \kgmm : Boxplots of min-max normalized log-likelihoods for the grid dataset.}}
\label{fig:grid-results}
\end{figure}

\clearpage
\begin{figure}[htb]
\begin{center}
\begin{tabular}{cc}
\rotatebox{90}{\kgmmseedingEGDEGC} & \includegraphics[width=.7\linewidth]{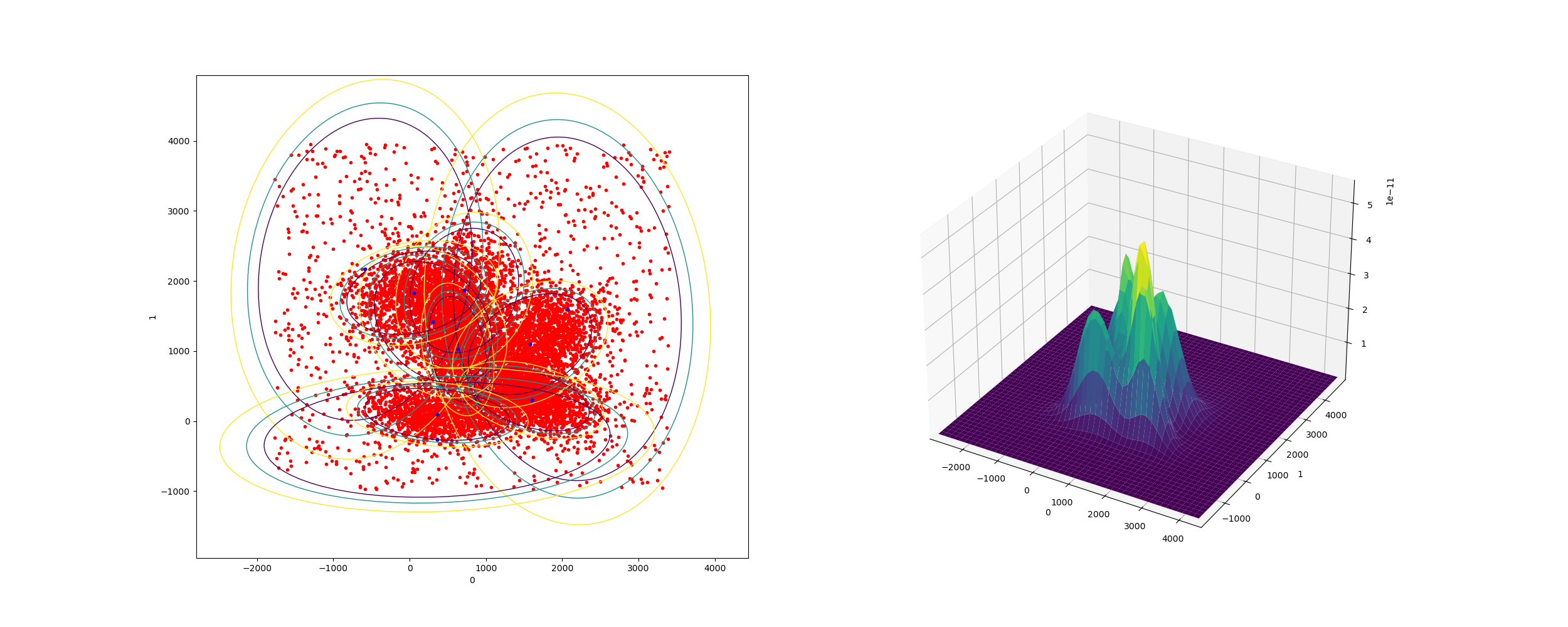}\\
\rotatebox{90}{\kgmmseedingEGDEGL} & \includegraphics[width=.7\linewidth]{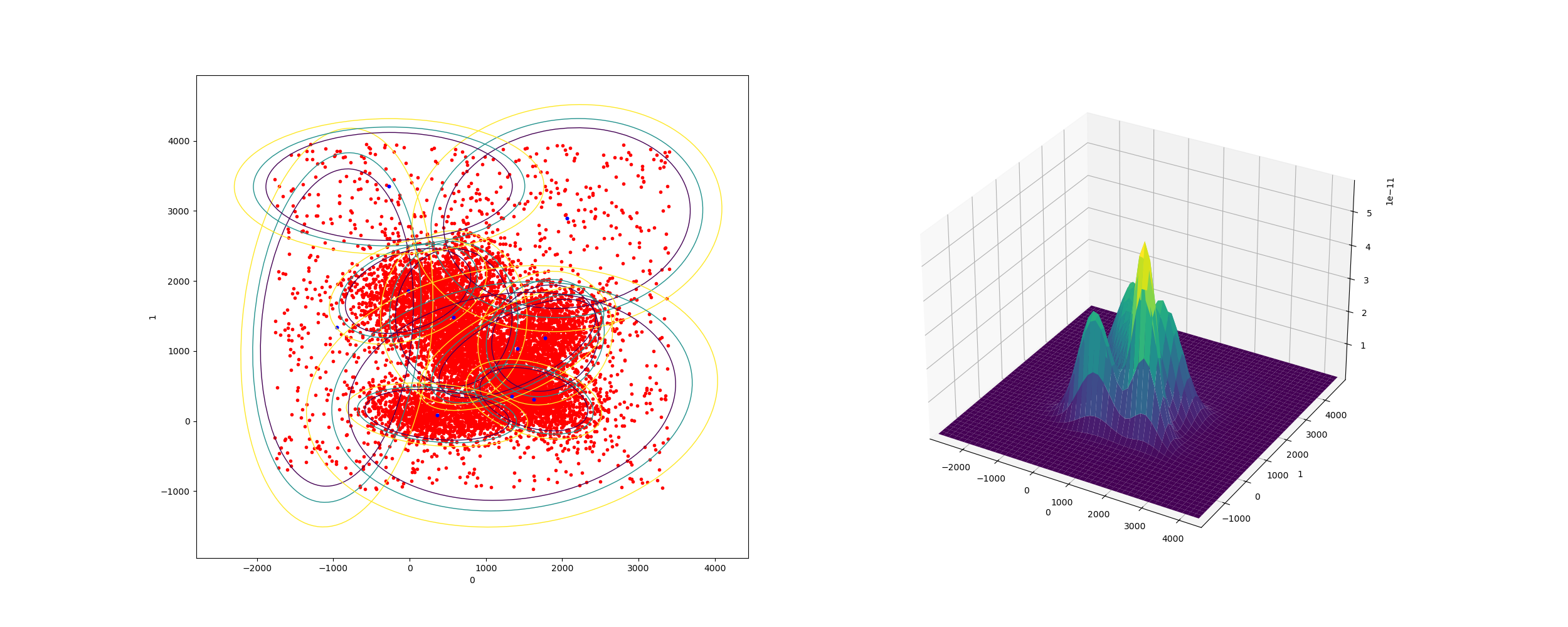}
\end{tabular}
\end{center}
\caption{{\bf \kgmm : Fitted models on a \textit{noisy-elliptical} dataset
with separation $s=0.5$.} Data is generated using a GMM with 10
components.  Each component is represented with a
blue dot for  the mean, and contour lines to represent the confidence
regions at [0.85, 0.90, 0.95] (darker lines corresponding to higher
confidence).  Likelihood regulated methods (bottom) are highly
sensitive to noise and consistently outperformed by SSE regulated
methods (top).}
\label{fig:noisy-elliptical-dot5}
\end{figure}

\begin{figure}[htb]
\begin{center}
\begin{tabular}{cc}
\rotatebox{90}{\kgmmseedingEGDEGC} & \includegraphics[width=.7\linewidth]{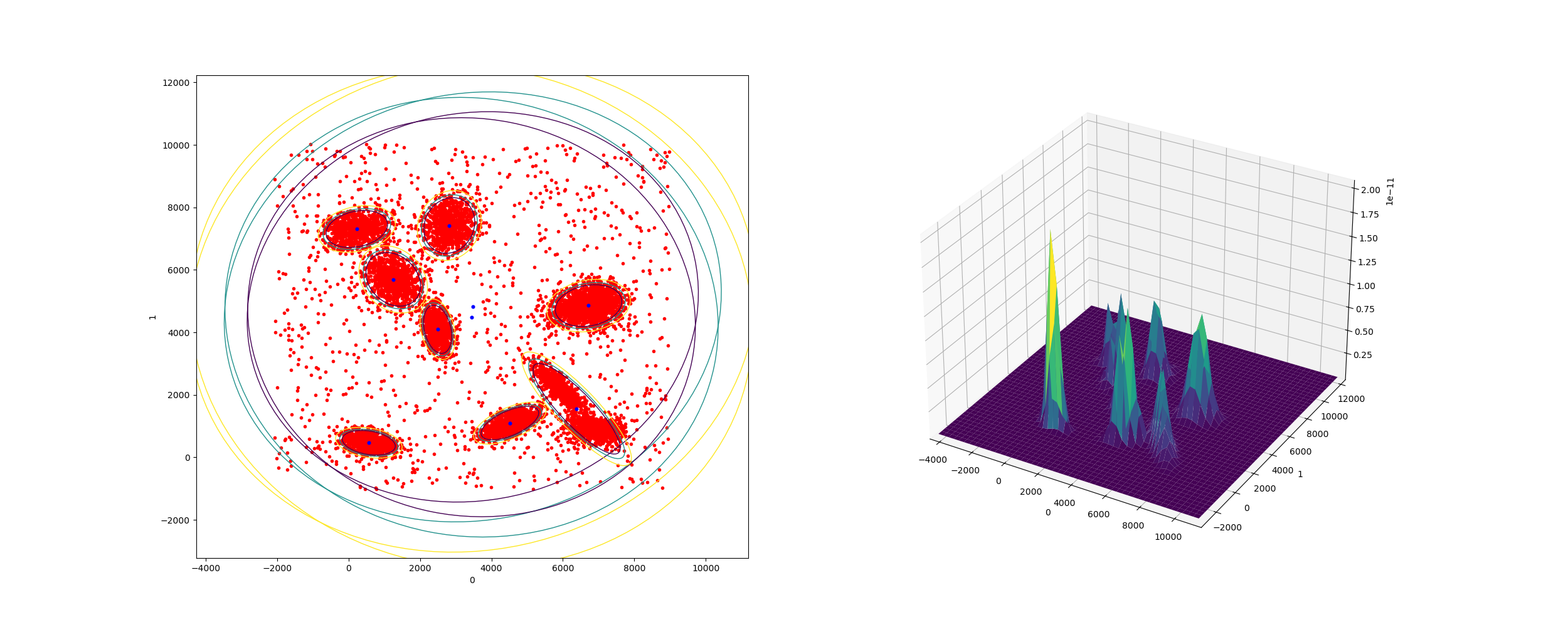}\\
\rotatebox{90}{\kgmmseedingEGDEGL} & \includegraphics[width=.7\linewidth]{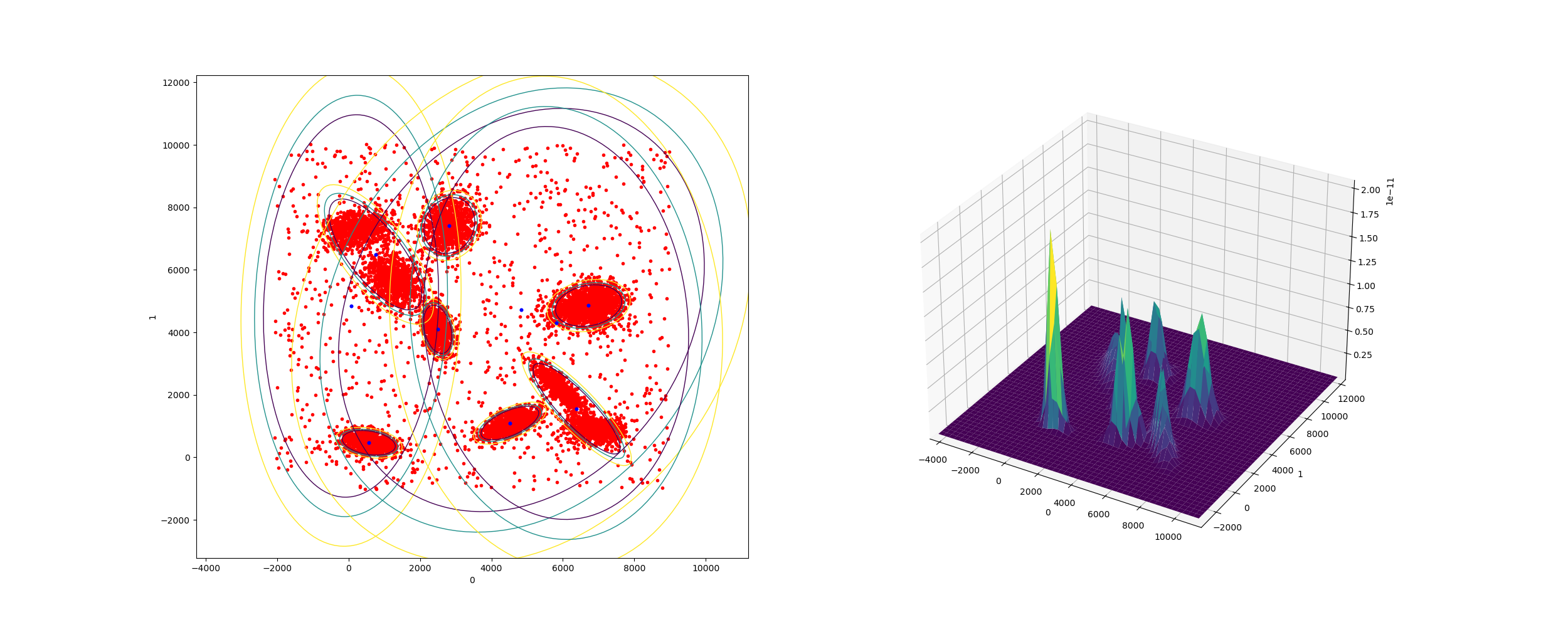}
\end{tabular}
\end{center}
\caption{{\bf \kgmm : Fitted models on a \textit{noisy-elliptical} dataset with separation $s=2$.} Data is generated using a GMM with 10 components. Conventions identical to Fig. \ref{fig:noisy-elliptical-dot5}.}
\label{fig:noisy-elliptical-sep2}
\end{figure}

\clearpage
\ifLONG
\else
\begin{figure}[htbp]
\centering
\includegraphics[width=.6\linewidth]{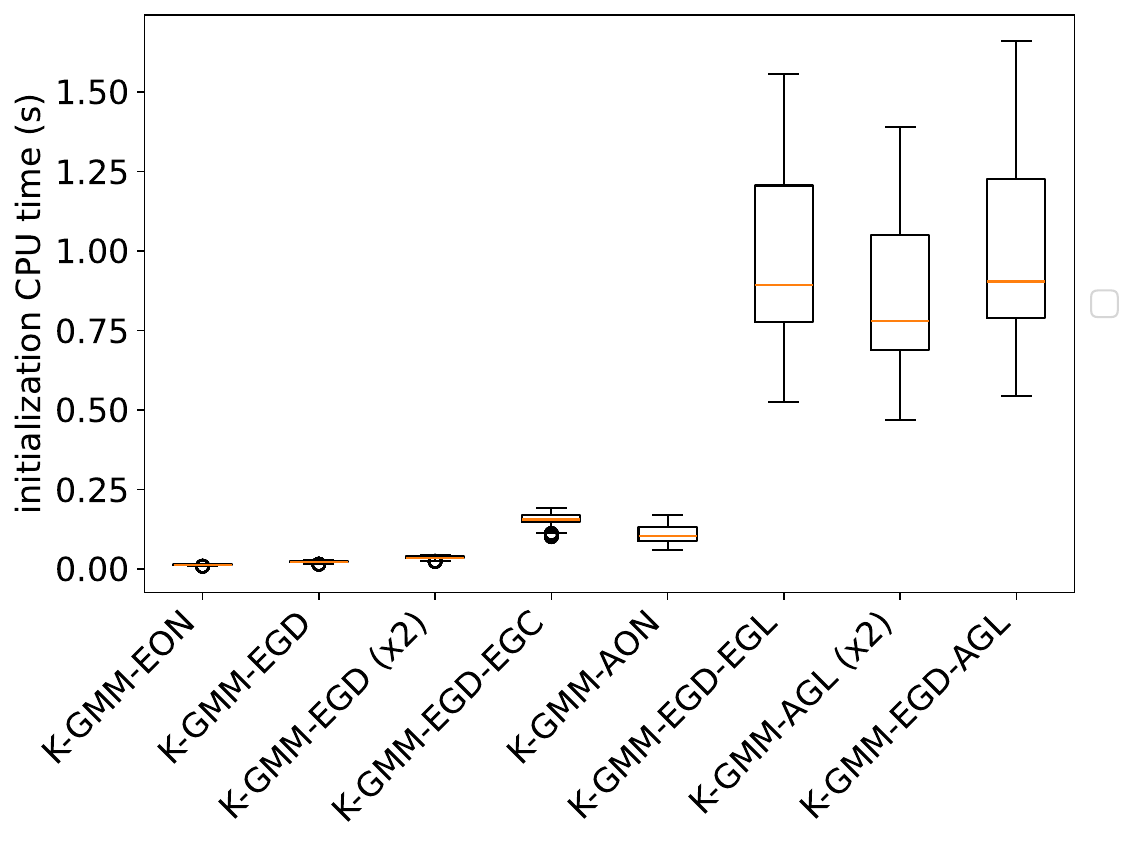}
\caption{{\bf \kgmm : average CPU time in seconds over all datasets for initialization only, when using each seeding method.}}
\label{fig:seeding-time-EM}
\end{figure}
\fi

\ifLONG
\else
\begin{figure}[htbp]
\centering
\includegraphics[width=.6\linewidth]{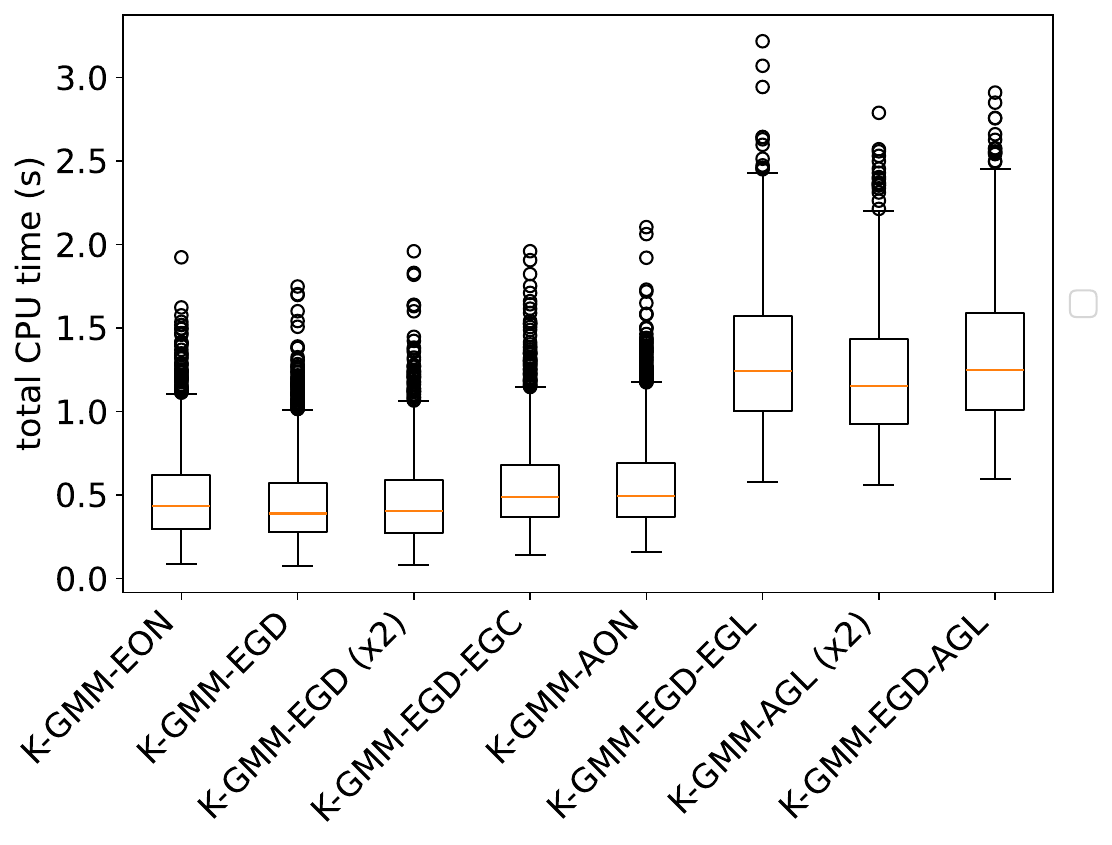}
\caption{{\bf \kgmm : average CPU time in seconds over all datasets for a full EM run, when using each seeding method.}}
\label{fig:total-time-EM}
\end{figure}
\fi

\begin{landscape}
\begin{table}
\tiny
\center
\begin{tabular}{|c|c|c|c|c|c|c|c|c|c|c|}
\hline
& \tiny EON & \tiny EGD & \tiny EGD (x2) & \tiny EGD-EGC & \tiny EGD-EGL & \tiny AGL (x2) & \tiny EGD-AGL & \tiny AON & \tiny GGD (x2) \\\hline
sphericalk10 & \tiny 0.51 & \tiny 0.78 & \tiny 0.83 & \tiny 0.91 & \tiny 0.92 & \tiny 0.92 & \tiny \bf{0.95} & \tiny 0.33 & \tiny 0.77\\
  & \tiny ± 0.24 & \tiny ± 0.18 & \tiny ± 0.16 & \tiny ± 0.15 & \tiny ± 0.13 & \tiny ± 0.11 & \tiny ± 0.10 & \tiny ± 0.23 & \tiny ± 0.22\\\hline
ellipticalk10 & \tiny 0.47 & \tiny 0.73 & \tiny 0.81 & \tiny \bf{0.93} & \tiny 0.85 & \tiny 0.89 & \tiny \bf{0.93} & \tiny 0.25 & \tiny 0.79\\
  & \tiny ± 0.22 & \tiny ± 0.18 & \tiny ± 0.17 & \tiny ± 0.14 & \tiny ± 0.20 & \tiny ± 0.14 & \tiny ± 0.14 & \tiny ± 0.20 & \tiny ± 0.16\\\hline
elliptical-difficultk10 & \tiny 0.50 & \tiny 0.74 & \tiny 0.78 & \tiny 0.82 & \tiny 0.88 & \tiny 0.87 & \tiny \bf{0.91} & \tiny 0.19 & \tiny 0.78\\
  & \tiny ± 0.23 & \tiny ± 0.20 & \tiny ± 0.20 & \tiny ± 0.24 & \tiny ± 0.18 & \tiny ± 0.15 & \tiny ± 0.14 & \tiny ± 0.16 & \tiny ± 0.26\\\hline
noisy-sphericalk10 & \tiny 0.32 & \tiny 0.73 & \tiny 0.79 & \tiny \bf{0.83} & \tiny 0.48 & \tiny 0.65 & \tiny 0.66 & \tiny 0.39 & \tiny 0.78\\
  & \tiny ± 0.21 & \tiny ± 0.20 & \tiny ± 0.22 & \tiny ± 0.23 & \tiny ± 0.29 & \tiny ± 0.24 & \tiny ± 0.24 & \tiny ± 0.20 & \tiny ± 0.24\\\hline
noisy-ellipticalk10 & \tiny 0.31 & \tiny 0.82 & \tiny 0.91 & \tiny \bf{0.94} & \tiny 0.19 & \tiny 0.45 & \tiny 0.50 & \tiny 0.46 & \tiny 0.87\\
  & \tiny ± 0.18 & \tiny ± 0.12 & \tiny ± 0.12 & \tiny ± 0.10 & \tiny ± 0.21 & \tiny ± 0.21 & \tiny ± 0.20 & \tiny ± 0.26 & \tiny ± 0.13\\\hline
noisy-elliptical-difficultk10 & \tiny 0.27 & \tiny 0.77 & \tiny 0.86 & \tiny 0.83 & \tiny 0.24 & \tiny 0.51 & \tiny 0.51 & \tiny 0.38 & \tiny \bf{0.89}\\
  & \tiny ± 0.22 & \tiny ± 0.16 & \tiny ± 0.15 & \tiny ± 0.20 & \tiny ± 0.24 & \tiny ± 0.20 & \tiny ± 0.22 & \tiny ± 0.25 & \tiny ± 0.19\\\hline
\end{tabular}
\caption{{\bf \kgmm : mean and standard deviation of min-max normalized log-likelihood for each seeding method}}
\label{tab:results_table_EM}
\end{table}
\end{landscape}

\clearpage
{\scriptsize
\tableofcontents
}

\end{document}